%% file: main.tex
\documentclass{article}

\usepackage[T1]{fontenc}
\usepackage{iclr2027_conference,times}
\usepackage{url}
\usepackage{graphicx}
\usepackage{booktabs}
\usepackage{longtable}
\usepackage{array}
\usepackage{amsmath}
\usepackage{amssymb}
\usepackage{mathtools}
\usepackage{amsthm}
\usepackage{xcolor}
\usepackage{float}
\usepackage{hyperref}
\hypersetup{hidelinks}

\usepackage{multirow}
\newcommand{\sd}[1]{{\scriptsize\,\(\pm\)\,#1}}

\newtheorem{definition}{Definition}[section]
\newtheorem{theorem}[definition]{Theorem}
\newtheorem{corollary}[definition]{Corollary}
\newtheorem{proposition}[definition]{Proposition}

\newcommand{\R}{\mathbb{R}}

\newcommand{\method}{\textsc{DistDNA}}
\newcommand{\distdnalinear}{DistDNA-Linear}
\newcommand{\distdnaexact}{DistDNA-RBF}

\input{data/completion_macros}

\usepackage{minitoc}
\title{\method: Language Model Fingerprinting under Stochastic Decoding}

\author{Lifan Hu, Qian Yu, Zhaomin Wu, Bingsheng He \\
Department of Computer Science, National University of Singapore \\
\texttt{\{lifan.hu, yuqian\}@u.nus.edu, \{zhaomin,dcsheb\}@nus.edu.sg}}

\iclrfinalcopy
\hypersetup{pdftitle={DistDNA: Language Model Fingerprinting under Stochastic Decoding},pdfauthor={Lifan Hu, Qian Yu, Zhaomin Wu, Bingsheng He}}
\begin{document}
\doparttoc
\faketableofcontents

\maketitle
\lhead{Preprint}

\begin{abstract}
\input{sections/abstract}
\end{abstract}

\section{Introduction}
\label{sec:introduction}
\input{sections/introduction}

\section{Related Work}
\label{sec:related-work}
\input{sections/related-work}

\section{Definition of \method}
\label{sec:problem}
\input{sections/problem-formulation}

\section{Extracting \method}
\label{sec:method}
\input{sections/method}

\section{Experiments}
\label{sec:experiments}
\input{sections/experiments}

\input{sections/results}

\section{Conclusion}
\label{sec:conclusion}
\input{sections/conclusion}

\section*{Reproducibility Statement}
The source package supports classifier and metric replay from separately archived frozen distance matrices and includes selected figure/table inputs. Recomputing fingerprints and calibration requires separately archived response and embedding banks.
Appendix~\ref{app:artifact-access} distinguishes these replay levels and lists the required inputs.

\section*{Ethical Statement}
We use public models for research under their published terms. Claims about theft, license violations or ancestry require corroborating evidence and an appeal process.

\section*{AI Use Statement}\label{apdx:llm-use}
\input{sections/llm_usage}

\bibliography{references}
\bibliographystyle{iclr2027_conference}

\clearpage
  \begingroup
  \renewcommand{\partname}{}
  \renewcommand{\thepart}{}
  \part{Appendix}
  \parttoc
  \endgroup
\appendix

\section{Theory and Proofs}\label{app:rfftrace-details}
\input{sections/appendix-definition}
\input{sections/appendix-theory}
\section{Experimental Details and Baseline Definitions}\label{app:experimental-details}
\input{sections/appendix-protocol}
\input{sections/appendix-uncertainty}
\section{Essential Ablations and Robustness}\label{app:supplementary}
\input{sections/appendix-motivation}
\input{sections/appendix-controls}

\input{sections/appendix-allocation}

\input{sections/appendix-parameters}

\input{sections/appendix-variance}
\input{sections/appendix-fresh}

\section{Limitations and Impact}\label{app:boundaries}

\subsection{Limitations}\label{app:limitations}
\input{sections/limitations}
\subsection{Broader Impact}\label{app:broader-impact}
\input{sections/broader-impact}
\section{Model and Artifact Inventory}\label{app:model-inventory}
\input{sections/appendix-inventory}

\clearpage
\section{Additional Identification Controls}\label{app:additional-controls}
\input{sections/appendix-additional-controls}
\input{sections/appendix-api-extension}

\end{document}

%% file: data/completion_macros.tex
\newcommand{\FreshConfidenceMargin}{11.91}

%% file: sections/abstract.tex
Behavioral fingerprints identify a large language model (LLM) from its responses to a shared set of prompts, enabling applications such as verifying whether a deployed model corresponds to the checkpoint it claims to be. Deployed models, however, are commonly served with stochastic decoding, under which the same model may produce different responses to an identical prompt, depending on an unknown temperature and top-$p$ setting. Existing representations such as LLM DNA provide limited treatment of this variation. A natural remedy is to characterize models by their response distributions, which, however, raises two challenges. First, each distribution must be estimated from only a small number of sampled responses. Second, a change in decoding setting can alter responses to the same extent as a change of checkpoint. Consequently, fingerprint distances alone provide an incomplete basis for model identity, and identification typically relies on a classifier trained jointly on fingerprints from many checkpoints. We propose \method{}, which represents a model by its prompt-conditional response distributions and compares models by promptwise maximum mean discrepancy (MMD). The model under examination is assigned to the known checkpoint whose reference fingerprints, collected at several decoding settings, lie nearest to its own, without training a classifier. We show that finite sampling adds variance-dependent terms to the expected squared fingerprint distance, with these terms decreasing as $O(1/R)$ for
equal response counts, where $R$ is the number of responses per prompt. Experiments across 115 model checkpoint entries demonstrate that \method{} improves identification accuracy over the strongest baseline, LLM DNA, by 10.36\%.

%% file: sections/introduction.tex
Learning representations of language models from their observable behavior has emerged as an important problem, with applications including API verification, model routing, and phylogenetic analysis. Behavioral approaches such as LLM DNA~\citep{wu2026llmdna} and PhyloLM~\citep{yax2025phylolm} characterize models through their responses to shared prompts, enabling comparison even when model parameters, tokenizers, or deployment details are unavailable. These representations have shown strong empirical effectiveness, achieving high accuracy in tasks such as predicting model relationships and lineage.

A practical example is verifying models served through third-party API relays. A service may claim to provide a specific checkpoint while hiding the actual backend. By querying the API with diagnostic prompts and comparing its responses against enrolled references, one can identify which checkpoint is actually deployed. This requires distinguishing even closely related checkpoints.

However, existing representation methods largely overlook stochastic decoding, despite its widespread use in modern large language model (LLM) systems. Under stochastic decoding, controlled by parameters such as temperature and nucleus sampling (\textit{top-p}), the same LLM can produce different responses to the same prompt~\citep{holtzman2020curious}. A single response is therefore only one sample from a prompt-conditional response distribution, rather than a complete representation of the response distribution. This motivates comparing models at the distribution level instead of through individual generated responses.

Fingerprinting models at the distribution level, however, presents two challenges. First, the response distribution of a model cannot be observed exhaustively. The space of possible responses to a prompt is too large to enumerate, and each query incurs a cost; the distribution must therefore be estimated from only a small number of sampled responses. Second, identification is difficult to achieve without a trained classifier. Since a change in temperature or top-$p$ can alter a model's responses as much as a change of checkpoint (Section~\ref{subsec:motivation}), distances between fingerprints do not directly determine model identity. Existing representations such as LLM DNA~\citep{wu2026llmdna} therefore need a trained classifier, albeit a lightweight one, which must be fit jointly on labeled fingerprints from many checkpoints, so that the decision for any one checkpoint depends on all other known checkpoints.

To address these challenges, we propose \method{}, which represents a model by its response distributions to a shared set of prompts. For the first challenge, \method{} draws repeated responses to each prompt and summarizes them by their mean embedding in a kernel feature space, which averages out sampling noise so that a small number of responses yields a stable fingerprint. For the second challenge, \method{} measures the distance between two models by the maximum mean discrepancy (MMD) with a Gaussian kernel, computed prompt by prompt. This distance compares the characteristic functions of the two response distributions across all frequencies~\citep{sriperumbudur2010hilbert}, thus reflecting differences in the full distribution rather than only in the mean response. The model under examination is then assigned to the known checkpoint whose reference fingerprints, collected at several decoding settings, lie nearest to its own, without training a classifier.

The contributions of this paper are threefold: (1) We propose \method{}, a behavioral fingerprint that characterizes a language model by its prompt-conditional response distributions, enabling identification by nearest enrolled reference without a trained classifier. (2) We prove that sampling noise inflates the estimated distance between two fingerprints, and that this bias vanishes at rate $O(1/R)$, where $R$ is the number of repeated responses per prompt. (3) Our experiments across 115 model checkpoint entries show that \method{} improves identification accuracy over the strongest baseline, LLM DNA, by 10.36\%.

%% file: sections/related-work.tex
\noindent\textbf{Behavioral representations.}
LLM-DNA constructs functional representations from shared probe responses for model comparison and downstream tasks, including relationship detection and routing~\citep{wu2026llmdna}. Extraction is training-free; downstream readouts may be trained. Lexical and morphosyntactic fingerprints also reveal model-family structure~\citep{mcgovern-etal-2025-large}. PhyloLM estimates prompt-conditioned output frequencies and uses distribution similarities to study model relationships and predict capabilities~\citep{yax2025phylolm}. The prior representations' native evaluations leave exact enrolled-checkpoint identification across hidden decoding shifts unestablished; \method{} targets this setting through repeated prompt-conditional kernel-mean estimation~\citep{gretton2012kernel} and separate enrollment references across settings.

\noindent\textbf{Direct model identification.}
LLMmap directly identifies model versions under unknown system prompts and stochastic sampling settings~\citep{pasquini2025llmmap}. Sensitive probes detect model tampering~\citep{bai-etal-2025-esf,hu2025resf}. Other fingerprints exploit optimized probes, counterfactual responses or seed behavior~\citep{hu2026llmprint,wu2026tcf,tong2026seedprints}. Watermarks and implanted fingerprints modify model behavior for attribution or ownership verification~\citep{kirchenbauer2023watermark,xu2024instructional,yamabe-etal-2025-mergeprint}; our protocol requires text access only. For settings where probe optimization or model modification is unavailable,
\method{} provides reusable fingerprints from fixed shared prompts, supporting
both nearest-reference matching and supervised identification with a support vector machine (SVM) under a common response budget.

Table~\ref{tab:prior-scope} summarizes representation choices and identification targets. \method{} combines repeated prompt-conditional observations with separate evaluation of distance-based and supervised readouts under hidden decoding settings.
\input{data/related_work_comparison}

%% file: data/related_work_comparison.tex
\begin{table}[!htbp]\centering\small
\setlength{\tabcolsep}{2pt}
\renewcommand{\arraystretch}{1}
\caption{\textbf{Fingerprinting methods differ in representation and identification target.} Checks denote explicit components or targets; --- means not established.}\label{tab:prior-scope}
\begin{tabular}{@{}>{\raggedright\arraybackslash}p{.13\linewidth}>{\raggedright\arraybackslash}p{.17\linewidth}>{\centering\arraybackslash}p{.09\linewidth}>{\centering\arraybackslash}p{.12\linewidth}>{\centering\arraybackslash}p{.11\linewidth}>{\centering\arraybackslash}p{.10\linewidth}>{\centering\arraybackslash}p{.09\linewidth}>{\raggedright\arraybackslash}p{.12\linewidth}@{}}
\toprule
\textbf{Method} & \textbf{Behavioral unit} & \textbf{Repeated draws} & \textbf{Response distribution} & \textbf{Decoding shift} & \textbf{Unknown setting} & \textbf{Exact ID} & \textbf{Readout}\\
\midrule
LLM-DNA & Prompt vectors & $\checkmark$ & $\checkmark$ & --- & --- & $\checkmark$ & Distance / classifier\\
PhyloLM & Output counts & $\checkmark$ & $\checkmark$ & --- & --- & --- & Distance\\
LLMmap & Query traces & --- & --- & $\checkmark$ & $\checkmark$ & $\checkmark^{*}$ & Classifier\\
McGovern & Text features & --- & $\checkmark$ & --- & --- & --- & Classifier\\
ESF & Token sets & --- & --- & $\checkmark$ & $\checkmark$ & --- & Set test\\
MergePrint & Implanted pairs & --- & --- & --- & --- & --- & Match\\
\midrule
\textbf{\method{}} & \textbf{Promptwise distributions} & $\checkmark$ & $\checkmark$ & $\checkmark$ & $\checkmark$ & $\checkmark$ & \textbf{Nearest / SVM}\\
\bottomrule
\end{tabular}
\par\smallskip\raggedright
Targets: model versions for LLMmap ($*$), relationships for PhyloLM, text detection for McGovern, tampering for ESF, ownership for MergePrint. Repeats are per prompt; nearest denotes nearest-reference matching.
\end{table}

%% file: sections/problem-formulation.tex
\noindent\textbf{Motivation: identifying models under decoding shift.}\label{subsec:motivation}
A useful model fingerprint must support identification when the query and
reference use different decoding settings. With a single response per prompt,
a change in the observed fingerprint can reflect a different checkpoint,
a different decoding setting, or simply another sample from the same
response distribution. How much decoding variation can a fingerprint tolerate
before exact checkpoint identification becomes unreliable?

\begin{figure}[htbp]
  \centering
  \includegraphics[width=0.9\linewidth]{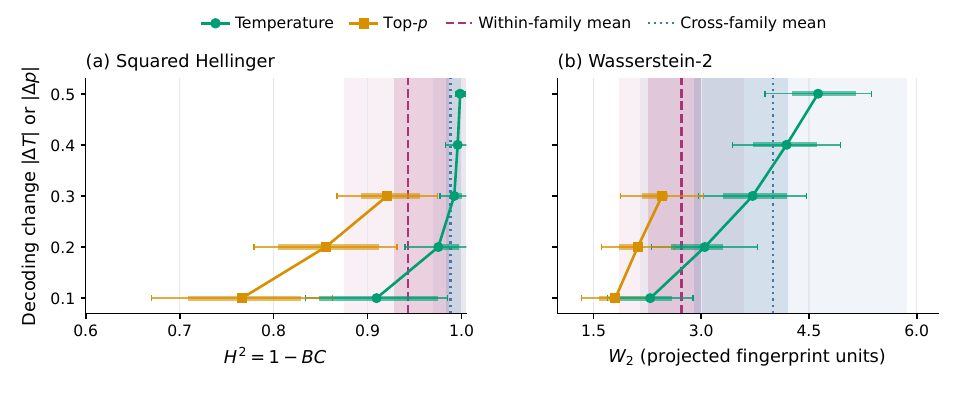}
  \caption{\textbf{Decoding shifts approach model-pair distance scales under two Gaussian metrics.} Original 32-seed fits across 41 checkpoints: (a) squared Hellinger $H^2$; (b) Wasserstein-2 $W_2$ in projected fingerprint units. Curves vary $T$ from $1$ at $p=1$, or $p$ from $1$ at $T=0.7$. Thin/thick bars show SD/IQR across checkpoints; dashed/dotted lines and light/dark bands show pooled within-/cross-family means and SD/IQR. The axes report metric-specific distributional separation.}
  \label{fig:atlas-motivation}
\end{figure}

\noindent\textbf{Observation 1: decoding shift reaches the scale of model differences.}
Figure~\ref{fig:atlas-motivation}(a) compares these effects on 41 checkpoints,
nine settings, and 32 generation seeds. We compare fitted Gaussian fingerprint
distributions using squared Hellinger distance, a measure of distributional
separation. At fixed top-$p=1$, changing $T$ from $1.0$ to $0.8$ gives a
mean distance of $0.975$, above the within-family reference of $0.943$.
At $T=0.7$ ($|\Delta T|=0.3$), it reaches $0.992$, comparable to the
cross-family reference of $0.988$. A fingerprint comparison must therefore
contend with decoding-induced changes as large as between-checkpoint changes.

\noindent\textbf{Observation 2: model-specific reference scales complicate identification.}
The complementary $W_2$ view in Figure~\ref{fig:atlas-motivation}(b)
shows the same tension: $T:1\to0.8$ exceeds the pooled within-family
reference ($3.049$ versus $2.727$), while $T:1\to0.7$ approaches the
cross-family reference ($3.715$ versus $4.003$). Across families,
within-family $H^2$ references span $0.913$--$0.996$
(Appendix Figure~\ref{fig:atlas-family-refs}). The direct identification tests below measure how these behavioral differences affect checkpoint recovery within the enrolled gallery.

\paragraph{Design implication.}
These observations motivate estimating prompt-conditional behavior from repeated responses and enrolling each checkpoint across decoding settings. Prompt alignment preserves the identity of each probe, while repeated sampling reduces within-prompt estimation error. The following theory characterizes this estimation mechanism; the identification experiments measure its practical effect under matched observation budgets.

A stochastic model produces a distribution of possible responses to each prompt. We define a behavioral distance between these conditional distributions and construct a population fingerprint whose Hilbert-space distance equals that distance. This separates the population object from the sampling algorithm in Section~\ref{sec:method}.

We separate representation from identification: a fingerprint estimates response-distribution differences; a readout uses these differences to predict an enrolled identity. The theory addresses the first question. The experiments evaluate the second.

\subsection{Prompt-Conditional Behavior}

Let $g=(m,\theta)$ denote checkpoint $m$ at decoding setting $\theta$. At prompt $x_i$, $Y_{g,i}\sim P_g(\cdot\mid x_i)$ and a frozen encoder with per-response unit normalization gives $Z_{g,i}=\phi(Y_{g,i})$. Fix prompts $S_t=(x_1,\ldots,x_t)$ and a measurable kernel feature map $\psi:\mathbb R^d\to\mathcal H$ with finite conditional second moments. Write $\mu_{g,i}=\mathbb E\psi(Z_{g,i})$ and define
\begin{equation}\label{eq:panel-distance}
d_{S_t,k}^2(g,h)=\frac1t\sum_{i=1}^t\|\mu_{g,i}-\mu_{h,i}\|_{\mathcal H}^2.
\end{equation}
Prompt alignment compares responses elicited by the same question. Linear $k$ measures mean differences; a characteristic RBF kernel distinguishes encoded response laws at the population level. The square root $d_{S_t,k}$ is a pseudometric on checkpoint--decoding pairs: different pairs can share the same encoded response laws on the selected prompts. It measures behavioral differences jointly induced by identity and decoding setting.

\subsection{Representation and Sampling Properties}
\begin{proposition}[Kernel-mean isometry]\label{prop:kernel-mean-isometry}
The population fingerprint
$\Gamma(g)=t^{-1/2}[\mu_{g,1};\ldots;\mu_{g,t}]\in\mathcal H^t$
satisfies
\begin{equation}\label{eq:population-fingerprint}
\|\Gamma(g)-\Gamma(h)\|_{\mathcal H^t}^2=d_{S_t,k}^2(g,h).
\end{equation}
\end{proposition}
The scaling makes squared distance an average over prompts. The proof expands the direct-sum norm; Appendix~\ref{app:membership-proof} gives the details. The RBF population fingerprint lives in a Hilbert space; the kernel evaluates its empirical distance directly from response samples. Finite-set compression is a separate existence result in Appendix~\ref{app:finite-existence}.

For $R_g$ responses per prompt, define the empirical fingerprint by
  \[
  \widehat\mu_{g,i}
  =\frac{1}{R_g}\sum_{r=1}^{R_g}\psi(z_{g,i,r}),
  \qquad
  \widehat\Gamma_g
  =t^{-1/2}[\widehat\mu_{g,1};\ldots;\widehat\mu_{g,t}].
  \]
  Each block estimates the response behavior for one fixed prompt.

\begin{theorem}[Finite-generation variance]\label{thm:repetition}
Condition on fixed prompts, fixed checkpoint--decoding pairs, independently
collected calibration data (separate responses used to set the kernel distance scale), and the resulting fixed 
square-integrable feature
map. For $R$ i.i.d. responses within each prompt, let $V_{g,i}=\mathbb E\|\psi(Z_{g,i})-\mu_{g,i}\|^2$. Then
\begin{equation}\label{eq:sample-variance}
\mathbb E\|\widehat\Gamma_g-\Gamma_g\|^2=\frac1{tR}\sum_i V_{g,i}.
\end{equation}
For independent query and reference arms,
\begin{equation}\label{eq:distance-bias}
\mathbb E\|\widehat\Gamma_q-\widehat\Gamma_r\|^2
=\|\Gamma_q-\Gamma_r\|^2+\frac1{tR_q}\sum_iV_{q,i}+\frac1{tR_r}\sum_iV_{r,i}.
\end{equation}
\end{theorem}

After centering, independence makes the within-prompt and cross-arm terms vanish in expectation (Appendix~\ref{app:rfftrace-details}). Repetition therefore reduces fixed-panel estimation MSE at rate $1/R$. Candidate-dependent variance also contributes to pairwise distance differences. The experiments evaluate identification and response allocation across prompt panels.

%% file: sections/method.tex
\begin{figure}[th]\centering
\includegraphics[width=\linewidth]{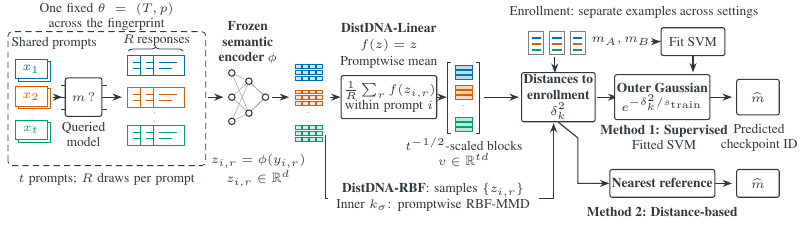}
\caption{\textbf{\method{} connects repeated response collection to setting-blind identification.} Prompt-aligned responses support linear means or RBF-MMD comparisons. Separate enrollment examples supply nearest references or train the SVM; an independent query enters prediction only. Colors identify prompts.}\label{fig:protocol}
\end{figure}

Figure~\ref{fig:protocol} summarizes the pipeline. Enrollment stores separate fingerprints across decoding settings. A query supplies repeated responses at one hidden setting; prompt-aligned comparisons support nearest-reference identification or a supervised SVM readout.

\subsection{Semantic Response Representation}\label{sec:semantic}
      Let $\phi$ denote the frozen semantic encoder followed by per-response unit 
      normalization, so that $z_{i,r}=\phi(y_{i,r})\in\mathbb R^d$ and $\|z_{i,r}\|
      _2=1$. Enrollment and queries share prompts and encoder conventions and use disjoint response draws. 
    Separate calibration prompts determine the inner bandwidth before query evaluation. This parameter 
    calibration sets a distance scale and the supervised outer-kernel scale is set separately from enrollment distances (Section~\ref{sec:classification}).

\subsection{Prompt-Conditional Distribution Comparison}\label{sec:estimating}

We use a positive bandwidth $\sigma$, with $\sigma^2$ set to the median
positive squared distance between calibration embeddings.
  
\noindent\textbf{\method{}-RBF.}
For $k_\sigma(z,w)=\exp[-\|z-w\|^2/(2\sigma^2)]$, compare empirical kernel means using
\begin{align}\label{eq:exact-mmd}
\widehat d_{\mathrm{RBF}}^2(q,r)=\frac1t\sum_i\Bigg[&
\frac1{R_q^2}\sum_{a,b=1}^{R_q}k_\sigma(z_{q,i,a},z_{q,i,b})\nonumber\\
+&\frac1{R_r^2}\sum_{a,b=1}^{R_r}k_\sigma(z_{r,i,a},z_{r,i,b})\nonumber\\
-&\frac2{R_qR_r}\sum_{a=1}^{R_q}\sum_{b=1}^{R_r}k_\sigma(z_{q,i,a},z_{r,i,b})\Bigg].
\end{align}
This V-statistic includes self-pairs and evaluates exact empirical kernel-mean distances from retained response samples. The comparison costs quadratic kernel work in the response count per prompt.

\noindent\textbf{\method{}-Linear.}
The linear construction averages within prompt and concatenates:
\begin{equation}\label{eq:mean-dna}
v_g=\frac1{\sqrt t}[\bar z_{g,1};\ldots;\bar z_{g,t}],\qquad
\bar z_{g,i}=\frac1R\sum_{r=1}^R z_{g,i,r}.
\end{equation}
Squared Euclidean distance between these vectors is empirical linear-kernel MMD. Prompt means and their concatenation retain their magnitudes. This construction is a simplification control for the response representation.

\subsection{Setting-Blind Identification}\label{sec:classification}
Both Linear and RBF constructions support the two readouts below: nearest-reference identification without identification-classifier training and supervised identification with a standard SVM. For nearest-reference identification, each checkpoint receives the minimum distance to any of its individual enrolled fingerprints; checkpoint scores rank distinct candidate IDs. Ties follow fixed enrollment order.
\begin{equation}
\widehat m(q)=\arg\min_{m\in\mathcal M}\;\min_{r\in\mathcal R_m}
\widehat d_k^{\,2}(q,r),
\end{equation}
where $\mathcal R_m$ contains the separately enrolled fingerprints of checkpoint $m$, and $k$ is the selected linear or RBF kernel.

  Supervised identification fits a separate multiclass SVM for each
  representation/encoder. Let
  $\delta_k^2(a,b)=\|\widehat\Gamma_a-\widehat\Gamma_b\|_{\mathcal H^t}^2$
  denote the empirical fingerprint distance. The common outer kernel is
\begin{equation}\label{eq:outer-kernel}
K(a,b)=\exp[-\delta_k^2(a,b)/s_{\mathrm{train}}],
\end{equation}
where $s_{\mathrm{train}}$ is the median positive off-diagonal enrollment squared distance. This converts fingerprint distances into classifier similarities. We fix $C=1$, use one-versus-one fitting and one-versus-rest ranking scores, and set the scale from enrollment only. The outer kernel is distinct from the inner response kernel; a genuine linear-kernel SVM is evaluated separately.

%% file: sections/experiments.tex
\subsection{Experimental Setting}\label{sec:experimental-setting}

\noindent\textbf{Datasets.}
Dolly instructions with empty context~\citep{databricks2023dolly} provide a category-balanced panel of 20 calibration prompts and 80 evaluation prompts. Main-experiment fingerprints use the first 20 evaluation prompts, disjoint from calibration. Calibration determines the inner bandwidth before query evaluation.

\noindent\textbf{LLMs.}
The main comparison contains 115 enrolled checkpoint entries, including base and instruction-tuned releases. The task identifies the exact enrolled checkpoint entry; every query identity belongs to the candidate set. Appendix~\ref{app:model-inventory} lists the entries and revisions. The original 30-checkpoint collection supplies supporting controls and the t-distributed stochastic neighbor embedding (t-SNE) illustration.

\noindent\textbf{Hyperparameters.}
Query settings form the full Cartesian grid $(T,p)\in\mathcal G\times\mathcal G$, where $\mathcal G=\{0.1,0.3,0.5,0.7,0.9,1.0\}$. Each fingerprint holds one hidden pair fixed across 20 prompts and four responses per prompt. This gives 4,140 physical queries and 331,200 fresh responses, with 80 responses per query. Each entry has eight separate enrollment fingerprints collected under $T\sim U(0.7,1.0]$ and independently $p\sim U(0.7,1.0]$: 920 references and 640 enrollment responses per entry. Enrollment, calibration and classifiers remain frozen across all 36 query cells.

\noindent\textbf{Baselines.}
LLM-DNA applies a shared 512-coordinate Gaussian projection (seed 42) to concatenated prompt means. PhyloLM-4char uses globally normalized prompt-specific four-character prefix frequencies. Both use the shared response banks and readouts; their task-specific adaptations are detailed in Appendix~\ref{app:baseline-constructions}. Random evaluates 1,000 fixed-seed uniform permutations of all 115 candidate IDs per physical query, reusing predictions across encoder views. The first ID supplies Top-1 accuracy and the F1 score, the harmonic mean of precision and recall; the first three distinct IDs supply Top-3 accuracy. Random has zero response-collection cost.

\noindent\textbf{Implementation.}
Pinned Beijing Academy of Artificial Intelligence (BAAI) General Embedding (BGE) and Masked and Permuted Pre-training for Language Understanding (MPNet) encoders produce unit-normalized 768-dimensional response embeddings. \method{}-Linear retains 15,360 coordinates; \method{}-RBF retains response samples for promptwise empirical MMD. Gaussian SVMs retain $C=1$ and their training-median scale. Nearest-reference matching ranks each checkpoint by its closest individual enrollment fingerprint under the declared representation distance. Both readouts use the same eight references per entry, frozen calibration and candidate set. All 682 empty query outputs are retained.

\noindent\textbf{Evaluation.}
For $k\in\{1,3\}$, model-macro Top-$k$ accuracy (hereafter denoted as Top-k) is
\begin{equation}
\mathrm{Top}\text{-}k
=\frac1M\sum_{m=1}^{M}\frac1{|\mathcal S|}\sum_{s\in\mathcal S}
\operatorname{Avg}_{e}\mathbf1\{m\in\operatorname{Top}_k(\mathbf a_{m,s,e})\},
\end{equation}
where $M=115$, $\mathcal S$ holds the 36 query settings, $e$ indexes encoder views, $\mathbf a_{m,s,e}$ are checkpoint scores, and $\operatorname{Top}_k$ ranks distinct IDs. Encoder outcomes are first averaged within each query; settings and checkpoints then receive equal weight, so every checkpoint counts equally. PhyloLM contributes one encoder-free view; each physical query remains one observation across encoder views. Macro F1 pools Top-1 predictions across settings and encoder views and averages per-checkpoint F1, $\mathrm{Macro\ F1}=\frac1M\sum_{m}2\mathrm{TP}_m/(2\mathrm{TP}_m+\mathrm{FP}_m+\mathrm{FN}_m)$, where TP, FP, and FN are true positives, false positives, and false negatives, and a zero denominator counts as zero. Tables report score $\pm$ sample SD across the 36 setting-level scores after encoder aggregation; central Macro F1 stays pooled, and its SD uses setting-specific pooled F1. These SDs quantify variation across decoding conditions. Random averages complete-trial metrics over 1,000 trials; its SD is across trial-averaged setting scores and reflects Monte Carlo variation.

%% file: sections/results.tex
\subsection{Main Identification Results}\label{sec:main-results}\label{sec:expanded}
Across 115 entries and 36 decoding settings (Table~\ref{tab:model-identification}), \method{}-RBF achieves 81.29\% Top-1 with nearest-reference matching and 89.93\% with the SVM, compared with LLM-DNA's 73.66\% and 84.20\%.
\method{}-Linear reaches 80.37\% and 89.42\%, respectively, and has the highest 
supervised Top-3. Both constructions improve on the evaluated external baselines under both readouts. The 
original thirty-checkpoint paired uncertainty analysis remains separate in Appendix~\ref{app:paired-uncertainty}. Sections~\ref{sec:readout} and \ref{sec:main-rejection} report the 
additional representation, rejection and native-chat results; Appendix~\ref{app:additional-controls} gives the 
detailed controls.
\input{data/main_continuous_svm_table}

\subsection{Representation and Readout Controls}\label{sec:readout}
On the thirty-checkpoint control, promptwise means with a linear-kernel SVM match the Gaussian SVM's 98.96\% Top-1. Simple means and a linear decision function suffice in this regime. \method{}-Linear retains 15,360 coordinates, while LLM-DNA compresses them to 512; Appendix~\ref{app:controls} compares the verified 128-, 512-, and 1024-coordinate constructions.

Across the SVM parameter grid, Top-1 varies by 0.20 percentage points for \method{}-RBF and 0.26 for Linear; RBF bandwidth changes yield 98.50--98.96\% (Appendix~\ref{app:parameters}).

\paragraph{Full-gallery representation controls.}
Bidirectional matched-coordinate controls across all 115 entries (4,140 queries) confirm that identification accuracy is governed by preserved gallery geometry and distributional moments rather than coordinate count (Appendix~F.3). Expanding LLM-DNA's gallery-blind Gaussian sketch (Appendix~B.2) $30\times$ from 512 to 15,360 coordinates adds only $+0.31$ percentage points on nearest-reference matching ($73.66\%$ to $73.97\%$). Conversely, compressing DistDNA to 512 coordinates (4~KiB) via unlabeled enrollment-span projection---realizing Corollary~A.2 with PCA for Linear and Nystr\"om for RBF using zero labels or extra queries---retains $80.31\%$ and $82.87\%$ nearest-reference accuracy ($86.79\%$ and $87.47\%$ with SVM) by preserving the rank-$\le 919$ reference manifold while filtering $>14{,}800$ orthogonal finite-sample noise axes, whereas replacing enrollment-span projection with an oblivious 512-d Gaussian sketch reverts to $\sim 73.6\%$. Similarly, while first-order means already separate $\sim 80\%$ of gallery queries so Linear and RBF tie across easy pairs ($+0.92$~pp overall), slicing the same 4,140 real queries shows that RBF lifts nearest-reference accuracy from $37.14\%$ to $41.00\%$ ($+3.86$~pp) on the $20\%$ most crowded rival queries ($n=828$), recovers $27.28\%$ overall and $42.68\%$ at high temperatures on mean-invariant real response residuals where Linear collapses to $0.87\%$ chance at any width, and improves equal-mean four-draw separation from $58.30\%$ to $91.60\%$ (Appendix~F.6). Uncompressed RBF stores $4\times$ Linear's payload; Appendix~F.4 details storage and CPU timings.

\subsection{Representation Geometry}\label{sec:geometry}
\begin{figure}[htb]\centering
\includegraphics[width=\linewidth]{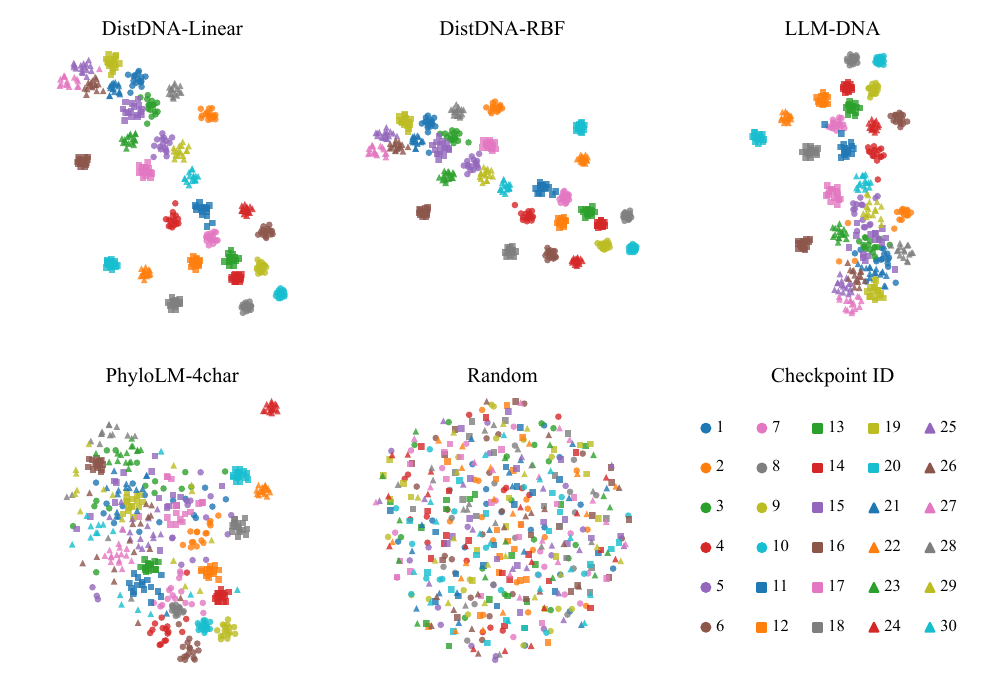}
\caption{Fingerprint geometry with BGE. Random is an output-space control; checkpoint IDs are listed in Appendix~\ref{app:visual-ids}.}\label{fig:main-tsne}
\end{figure}
\noindent\textbf{Geometry.} Figure~\ref{fig:main-tsne} shows checkpoint-specific groups across 480 queries from the original thirty-checkpoint collection. Random projects candidate-rank vectors; representation panels are descriptive, with predictive performance measured in the original space.

\begin{figure}[htb]\centering
\includegraphics[width=\linewidth]{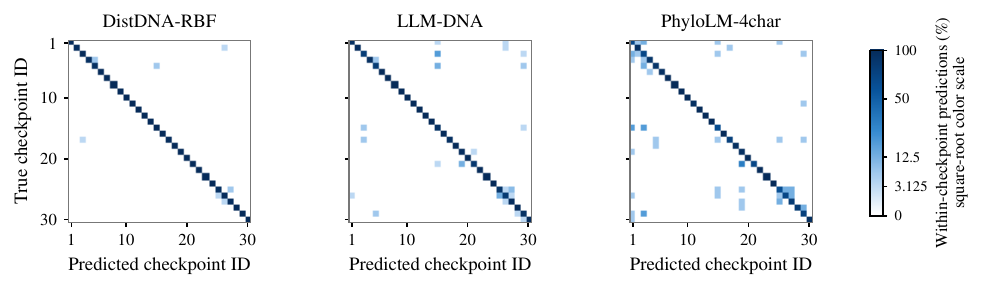}
\caption{SVM confusion matrices. Rows are true checkpoint IDs and columns are predictions; panels share row normalization and a square-root color scale.}\label{fig:main-svm}
\end{figure}
\noindent\textbf{Checkpoint confusions.} Figure~\ref{fig:main-svm} shows fewer RBF SVM errors than LLM-DNA and PhyloLM-4char on these 480 queries, pooling encoder views where applicable.

\subsection{Sampling and Enrollment Requirements}\label{sec:observations}
\begin{figure}[htb]\centering
\includegraphics[width=0.9\linewidth]{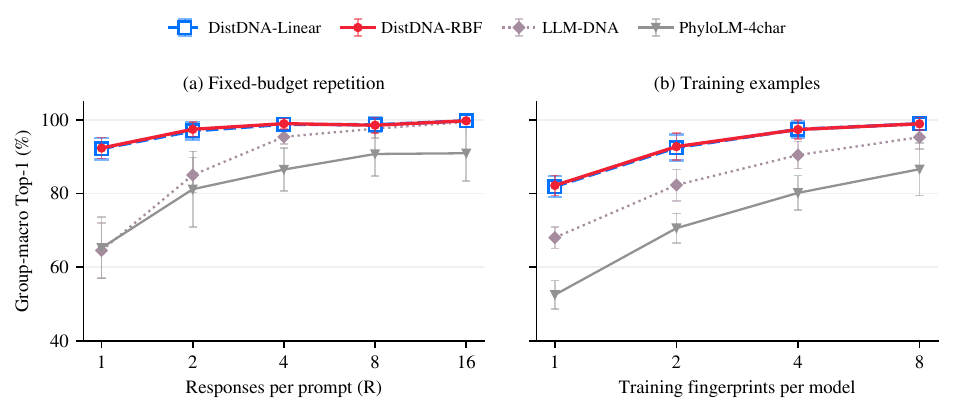}
\caption{SVM identification versus (a) repetitions at fixed response budget and (b) enrollment fingerprints per model. Bars show sample SD across sixteen held-out settings.}\label{fig:main-ablations}
\end{figure}
\noindent\textbf{(a) Repetition improves fixed-budget identification.} In Figure~\ref{fig:main-ablations}(a), \method{}-RBF rises from 92.37\% at $80\times1$ to 99.77\% at $5\times16$, with 80 query and 640 enrollment responses per checkpoint. The trajectory includes a local decrease from $20\times4$ to $10\times8$; the primary allocation remains $20\times4$.

\noindent\textbf{(b) Broader enrollment improves identification.} One versus eight references raises RBF SVM Top-1 from 82.21\% to 98.96\%, increasing setting coverage and response cost together. This curve uses the primary cache; allocation uses fresh draws (Appendix~\ref{app:geometry}).
Simultaneous 95\% lower bounds support 80\% accuracy at 40 query responses in both studies, and 90\% at 80 responses in the primary-cache study; their comparison families differ (Appendices~\ref{app:allocation} and~\ref{app:fresh-completion}).

\subsection{Decoding Robustness}\label{sec:robustness}
\begin{figure}[!htb]\centering
\includegraphics[width=\linewidth]{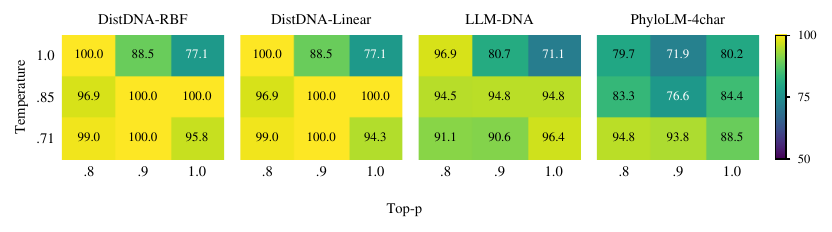}
\caption{Decoding robustness with fixed SVMs. Group-macro Top-1 (\%) on the 30-checkpoint study; all panels share the accuracy scale.}\label{fig:completed-decoding}
\end{figure}
\noindent\textbf{\method{} leads across most tested settings.} In Figure~\ref{fig:completed-decoding}, the leftmost panel, \method{}-RBF, has the highest nine-cell average (95.25\%), ahead of LLM-DNA (90.10\%) and PhyloLM-4char (83.68\%). It ties or exceeds both external baselines in seven of nine cells; \method{}-Linear closely matches it.

\noindent\textbf{Temperature interacts with nucleus filtering.} At $p=0.8$, \method{}-RBF remains above 96\% across the tested temperatures and reaches 100\% at $T=1$. At $T=1$, increasing $p$ to 0.9 and 1.0 reduces accuracy to 88.54\% and 77.08\%, respectively. Identification is strongest at $p=0.8$ and degrades as $p$ increases at $T=1$.

This separate thirty-checkpoint study crosses $T\in\{0.71,0.85,1.0\}$ with $p\in\{0.8,0.9,1.0\}$, using 80 responses/query and fixed enrollment, calibration, and SVMs. Greedy decoding is evaluated separately with 20 responses/query and yields 85.42\% for \method{}-RBF.

\input{sections/main-additional-results}

%% file: data/main_continuous_svm_table.tex
\begin{table*}[htbp]\centering\small
\setlength{\tabcolsep}{3.5pt}
\caption{\textbf{Identification across 115 enrolled entries and 36 decoding settings.} Model-macro accuracy and pooled Macro F1; score $\pm$ descriptive SD across grid cells. Bold and underlining indicate the best and second-best distinct scores within each readout and metric.}\label{tab:model-identification}
\begin{tabular}{l*{6}{c}}\toprule
\multirow{2.5}{*}{\textbf{Method}} & \multicolumn{3}{c}{\textbf{Supervised (SVM)}} & \multicolumn{3}{c}{\textbf{Nearest Reference}}\\
\cmidrule(lr){2-4}\cmidrule(lr){5-7}
 & \textbf{Top-1 (\%)} & \textbf{Top-3 (\%)} & \textbf{Macro F1} & \textbf{Top-1 (\%)} & \textbf{Top-3 (\%)} & \textbf{Macro F1}\\
\midrule
Random & N/A & N/A & N/A & 0.87\sd{0.02} & 2.61\sd{0.05} & 0.0086\sd{0.0002}\\
\midrule
LLM-DNA & 84.20\sd{7.03} & 93.61\sd{4.31} & 0.8422\sd{0.0756} & 73.66\sd{5.26} & 86.07\sd{5.37} & 0.7259\sd{0.0601}\\
PhyloLM-4char & 80.39\sd{5.98} & 93.50\sd{3.96} & 0.8015\sd{0.0636} & 61.28\sd{3.51} & 75.80\sd{4.32} & 0.5814\sd{0.0397}\\
LLMmap$^\dagger$ & 82.52\sd{4.52} & 93.12\sd{3.52} & 0.8251\sd{0.0563} & 63.55\sd{19.85} & 76.88\sd{17.63} & 0.6246\sd{0.2187}\\
\method{}-Linear & \underline{89.42}\sd{6.26} & \textbf{96.39}\sd{3.21} & \underline{0.8921}\sd{0.0722} & \underline{80.37}\sd{7.04} & \underline{91.51}\sd{4.64} & \underline{0.8002}\sd{0.0769}\\
\method{}-RBF & \textbf{89.93}\sd{5.71} & \underline{96.34}\sd{3.39} & \textbf{0.8978}\sd{0.0660} & \textbf{81.29}\sd{6.46} & \textbf{91.87}\sd{4.37} & \textbf{0.8087}\sd{0.0699}\\
\bottomrule\end{tabular}
\end{table*}

%% file: sections/main-additional-results.tex
\subsection{Generalizability Validation}\label{sec:main-validation}
\paragraph{Identification with rejection.}\label{sec:main-rejection}
An optional learned gate improves both unknown rejection and correct acceptance over a distance threshold. At the 95\% known-acceptance target, RBF unknown acceptance falls from 83.50\% to 52.17\%, while known rejection falls from 4.96\% to 3.79\%. Identity matching remains classifier-free. Appendix~\ref{app:trained-rejection-gate} gives the gate, other operating points and the official LLMmap detector comparison.

\paragraph{Native chat and reasoning.}\label{sec:main-native}
The new 86-service gallery covers six families and nearby versions. RBF reaches 86.82\% nearest-reference accuracy overall and 83.33\% on the 64-service reasoning-request cohort against all 86 candidates, with reference/query modes matched. Appendix~\ref{app:native-expanded} reports both constructions, family coverage and the reasoning-output audit; Appendix F.12 adds matched Dolly, BIG-Bench Hard (BBH), and MMLU-Pro comparisons.

%% file: sections/conclusion.tex
This paper studied exact checkpoint identification under hidden stochastic decoding settings and proposed \method{}, which represents models through prompt-conditional response distributions. We established its finite-sampling properties and evaluated both nearest-reference and supervised identification across 115 checkpoint entries and diverse decoding settings. Our results show that repeated response sampling and broader enrollment improve identification, supporting behavioral fingerprinting as a practical tool for black-box model verification.

%% file: sections/llm_usage.tex
Generative AI tools were used to assist with implementation verification. GPT-6 Astra was used to verify theoretical derivations and proofs after they had been developed by the authors; it was not used to formulate the theoretical claims or proofs. Claude Opus 5 was used for language editing and manuscript polishing. All AI-assisted outputs were reviewed and verified by the authors, who take responsibility for the final content of the paper.

%% file: sections/appendix-definition.tex
\subsection{Population Distance and Finite-Set Existence}\label{app:finite-existence}
For a prompt distribution $\nu$, assume integrable conditional feature second moments and define
\begin{equation}\label{eq:distributional-functional-distance}
d_{\nu,k}^2(g,h)=\mathbb E_{x\sim\nu}\|\mu_{g,x}-\mu_{h,x}\|_{\mathcal H}^2.
\end{equation}
The prompt-panel distance in Eq.~\eqref{eq:panel-distance} is its fixed-panel counterpart. With a characteristic kernel, zero population distance establishes equality of encoded conditional laws for almost every prompt; weight identity and training history require separate evidence.

\begin{definition}[Finite-dimensional distance preservation]\label{def:distributional-dna}
For fixed prompts, encoder and kernel and a finite behavior set $\mathcal G_K$, a finite representation $\tau:\mathcal G_K\to\mathbb R^L$ preserves the declared distance if
\begin{equation}\label{eq:distributional-dna-criterion}
c_1d_{S_t,k}(g,h)\leq\|\tau_g-\tau_h\|_2\leq c_2d_{S_t,k}(g,h)
\end{equation}
for positive constants $c_1\leq c_2$ and every pair in $\mathcal G_K$.
\end{definition}
\begin{corollary}[Finite-set compression]\label{cor:finite-distdna}
For $K\geq2$ and $0<\epsilon<1$, there exists a linear map on the span of the $K$ population fingerprints into $L=O(\epsilon^{-2}\log K)$ coordinates satisfying
\begin{equation}\label{eq:finite-distdna}
\sqrt{1-\epsilon}\,d_{S_t,k}(g,h)\leq\|A\Gamma(g)-A\Gamma(h)\|_2
\leq\sqrt{1+\epsilon}\,d_{S_t,k}(g,h).
\end{equation}
\end{corollary}
\begin{proof}
The finite collection spans a finite-dimensional Euclidean subspace of the Hilbert direct sum. Apply the Johnson--Lindenstrauss lemma to those $K$ points and use Proposition~\ref{prop:kernel-mean-isometry}~\citep{johnson1984extensions,larsen2017optimality}.
\end{proof}
Primary uncompressed evaluations compute Linear distances in $\mathbb{R}^{td}$ and empirical RBF distances directly from retained response samples, whereas Appendix~F.3 empirically realizes Corollary~A.2 at $L=512$ by projecting fingerprints onto the unlabeled span of the enrolled references via PCA (Linear) and kernel Nystr\"om (RBF). Finite-dimensional span compression and the infinite-dimensional population embedding thus play complementary operational roles.

%% file: sections/appendix-theory.tex
\subsection{Proof of Proposition~\ref{prop:kernel-mean-isometry}}\label{app:membership-proof}
The direct-sum inner product gives
\[
\|t^{-1/2}[\mu_{g,1}-\mu_{h,1};\ldots;\mu_{g,t}-\mu_{h,t}]\|^2
=t^{-1}\sum_i\|\mu_{g,i}-\mu_{h,i}\|^2.
\]
This is Equation~\eqref{eq:panel-distance}. The direct-sum norm adds prompt-block contributions independently. The identity applies to ordinary mean vectors and RBF kernel means alike.

\subsection{Proof of Theorem~\ref{thm:repetition}}\label{app:variance-proof}
Condition on the prompt panel, independently admissible calibration and fixed feature map. Let $\xi_{i,r}=\psi(Z_{i,r})-\mu_i$. These variables are centered and independent across draws within a prompt, so
\[
\mathbb E\left\|R^{-1}\sum_r\xi_{i,r}\right\|^2
=R^{-2}\sum_r\mathbb E\|\xi_{i,r}\|^2=V_i/R.
\]
Cross-draw inner products vanish by independence and centering. Averaging the prompt-block squared norms proves Eq.~\eqref{eq:sample-variance}; the direct-sum identity permits dependence across prompt blocks for this expectation.

For the second statement, write the empirical difference as the population difference plus $\epsilon_q-\epsilon_r$. Both errors have zero mean, and independent arms give $\mathbb E\langle\epsilon_q,\epsilon_r\rangle=0$. Therefore its expected squared norm is the population squared distance plus the two error MSEs. Applying the first identity with $R_q$ and $R_r$ proves Eq.~\eqref{eq:distance-bias}.

Unit-normalized linear features have $V_i\leq1$; the RBF feature map also has squared norm $k(z,z)=1$, so $V_i\leq1$. Markov's inequality consequently gives $\Pr(\|\widehat\Gamma-\Gamma\|\geq\varepsilon)\leq1/(R\varepsilon^2)$ for either retained construction. This conservative bound controls fingerprint-estimation error under the stated fixed-map and within-prompt i.i.d. conditions. Classification accuracy and sample-dependent normalization require separate analyses.

\subsection{Exact Empirical Kernel Distance}
The reproducing identity $\langle k(z,\cdot),k(w,\cdot)\rangle=k(z,w)$ expands the squared difference of empirical means into Eq.~\eqref{eq:exact-mmd}. All within-arm pairs, including self-pairs, remain; the cross denominator is $R_qR_r$. The V-statistic is nonnegative up to numerical roundoff. For $R_g\geq2$, define the unbiased within-prompt feature variance
  \[
  \widehat V_{g,i}
  =\frac{1}{R_g-1}
  \sum_{a=1}^{R_g}
  \|\psi(z_{g,i,a})-\widehat\mu_{g,i}\|_{\mathcal H}^2.
  \]
  The variance-corrected estimator is
  \[
  \widehat d_U^2
  =\widehat d_V^2
  -\frac1t\sum_{i=1}^t
  \left(
  \frac{\widehat V_{q,i}}{R_q}
  +\frac{\widehat V_{r,i}}{R_r}
  \right).
  \]
  This equals the unbiased MMD U-statistic for independent arms with
  $R_q,R_r\geq2$. Both primary readouts use the V-statistic.

\subsection{A Sufficient Margin for Nearest-Reference Identification}
If each estimated squared candidate distance is within $\eta$ of its population counterpart and the nearest correct candidate is separated from every incorrect one by more than $2\eta$, nearest-reference ranking remains correct. Subtracting the two estimation errors proves this sufficient condition. The condition applies to nearest-reference retrieval when a positive population margin persists under decoding shift; SVM prediction requires a separate analysis. At fixed total response budget, changing the prompt set changes both the population margin and the promptwise variances.

%% file: sections/appendix-protocol.tex
\subsection{Primary Continuous-Setting Collection}\label{app:continuous-settings}
The original thirty-checkpoint controls report Top-1/Top-3 by averaging queries and encoder views within checkpoint, checkpoints within each of eight reporting groups, and groups equally. These retained scores differ in aggregation from the equal-checkpoint 115-entry evaluation in Section~\ref{sec:experimental-setting}.

The original collection identifies one of 30 enrolled checkpoints from a query fingerprint with one fixed decoding setting hidden from the readout. Eight enrollment pairs and sixteen independent test pairs are sampled using independent children of NumPy SeedSequence 2026091519: $T\sim U(0.7,1.0]$ and $p\sim U(0.7,1.0]$. Pairs are shared across checkpoints. Every fingerprint uses 20 prompts and four responses per prompt; enrollment has 240 fingerprints and testing has 480. Calibration uses 20 disjoint prompts at the first enrollment pair.

\paragraph{Prompts and generation.}
The panel uses fresh Dolly instructions with empty context and dataset-provided reference answers of 40--180 words. Five categories contribute equally to the underlying 100-prompt panel. Deterministic hash ordering assigns 20 calibration and 80 evaluation prompts; the first 20 interleaved evaluation prompts are used here. Exact matches to prior configuration-manifest prompts and token-set Jaccard overlap of at least 0.90 are excluded. The collection manifest stores every prompt and hash.
All primary generations use single-precision (float32) inference and plain-text prompts, without native chat templates. The instruction suffix requests three to six complete sentences without repeating the question. Generation is capped at 128 new tokens and stops at four complete sentences, eight nonempty lines, or 72 whitespace-delimited words. Exactly one attempt is made; all 50 empty strings remain in the data and receive the ordinary encoder or empty-prefix treatment. The recorded collection runtime uses Torch 2.6.0+cu124 and Transformers 5.16.1 on NVIDIA A100-SXM4 graphics processing units (GPUs). Record seeds are fixed independently of worker assignment; numerical outputs may vary across runtimes.
\paragraph{Representations and fitting.}
BGE-base-en-v1.5 and all-mpnet-base-v2 use pinned revisions \texttt{a5beb1e3} and \texttt{e8c3b32e}; full identifiers are in the analysis lock. Each 768-coordinate response is unit normalized. \method{}-Linear preserves the scale of the prompt means and concatenated fingerprints. \method{}-RBF uses the nonnegative empirical MMD V-statistic, including self-pairs. The inner bandwidth is the median of at most 100,000 calibration distances, with the same seed. The outer SVM uses only the median positive off-diagonal training squared distance, $C=1$, precomputed Gaussian kernels, one-versus-one fitting, one-versus-rest decision scores and \texttt{break\_ties=True}. Repeated encoder evaluations share the same physical queries.

\paragraph{Metrics and statistical analysis.}
Top-1 and Top-3 first average encoder and setting measurements within model, models within group, and the eight reporting groups equally. Macro F1 pools the two encoders' predictions, computes F1 for each of the 30 model classes, and averages classes. PhyloLM-4char is encoder-free and is evaluated once. The collection protocol was frozen before generation; analysis code was frozen after collection and before identification scoring. The central scores are accompanied by descriptive sample SD across the sixteen held-out settings, as defined in Section~\ref{sec:experimental-setting}. Appendix~\ref{app:paired-uncertainty} adds retrospective paired uncertainty without changing these fits. The setting prior specifies the distribution used for experimental evaluation.

\input{sections/appendix-baselines}

\subsection{Artifact Availability}
The source package includes classifier and metric replay code, fixed random-English probes, and selected table and figure inputs. Replay additionally requires external frozen enrollment/query distance matrices, saved query predictions, and measured Random trials. Given the required external inputs, the replay refits the common SVM from frozen distances and verifies predictions, distinct-ID ranks, aggregation and table values without new collection. Recomputing representations and calibration additionally requires separately stored response/embedding banks and calibration inputs. Collection replay requires pinned model revisions, prompts, seeds and runtime conventions.\label{app:artifact-access}

%% file: sections/appendix-baselines.tex
\subsection{Baseline Construction and Readouts}\label{app:baseline-constructions}\label{app:baseline-source-validation}\label{app:phylolm-adaptation}
LLM-DNA projects concatenated prompt means with a shared double-precision (float64) Gaussian matrix generated by the 64-bit permuted congruential generator (PCG64), with entry variance $1/L$. The primary width is $L=512$ and projection seed 42; 128-coordinate and seed 43/44 controls are supplementary. Projection commutes with averaging. Response embeddings are normalized before averaging; prompt means and their concatenation retain the resulting scale.
The common readouts use squared Euclidean geometry, distinct from the imported workbook's cosine retrieval.

PhyloLM-4char counts the first four Unicode characters separately within each prompt, concatenates frequency vectors and globally normalizes them. Its squared distance is $2(1-S_{\mathrm{Nei}})$, where $S_{\mathrm{Nei}}$ is cosine similarity of the concatenated histograms. Four samples and the shared instruction prompts differ from its native contexts and sample sweep. The comparison matches response counts; token counts and storage costs remain method-dependent.

\paragraph{Measured Random.}
PCG64 seeds 2026092100--2026093099 each generate a uniform permutation of all thirty candidate IDs for every physical query. The same permutation is reused across encoder views and readout comparisons. Its first ID supplies Top-1 and F1; the first three distinct IDs supply Top-3. Central metrics average complete trials; the displayed SD measures variation across sixteen trial-averaged setting metrics.
Trial predictions are preserved in the accompanying metric-replay inputs.

\paragraph{Nearest reference.}\label{app:readout-protocol}
Each representation uses the same 240 enrollment fingerprints and 480 physical queries as the SVM. A checkpoint receives its minimum distance over eight enrolled fingerprints; distinct checkpoint minima determine the ranking. Exact ties use the first minimizing record in setting-major, then frozen manifest-model order. Every query is ranked against the full enrolled candidate set. The Top-1 predictions reproduce the prior matched-bank diagnostic. The 115-entry main comparison uses the counts and aggregation specified in Section~\ref{sec:experimental-setting}. Nearest-reference assignment ranks checkpoints by fingerprint distance; LLM-DNA's organization heuristic predicts binary relationships, and greedy decoding selects response tokens.

%% file: sections/appendix-uncertainty.tex
\subsection{Retrospective Paired Primary Uncertainty}\label{app:paired-uncertainty}
The estimand is the Top-1 difference between each retained \method{} construction and each evaluated baseline. Encoder correctness is averaged within checkpoint and setting before aggregation. The retrospective protocol uses 10,000 crossed bootstrap replicates, resampling sixteen shared test settings and checkpoints within each of the eight fixed reporting groups at its original size. Identical setting and checkpoint multiplicities apply to every method. Classifiers remain fixed; resampling operates on shared settings and checkpoints while preserving dependence among encoder views and overlapping enrollment windows.

The simultaneous margin is the 95th percentile of the maximum absolute centered bootstrap deviation. Both retained constructions have a 3.65-percentage-point gain over LLM-DNA with simultaneous 95\% interval $[-2.44,9.73]$, and a 12.34-point gain over PhyloLM-4char with interval $[6.25,18.42]$. The simultaneous interval for the LLM-DNA contrast includes zero.

%% file: sections/appendix-motivation.tex
\subsection{Seed and Sample-Size Checks for the Motivation Diagnostic}
\label{app:motivation-diagnostic}

\paragraph{Purpose and setup.}
This section supplies the estimation protocol and replication checks for
Figure~\ref{fig:atlas-motivation}. The historical panel contains 41 checkpoints,
100 shared random-Chinese probes, and nine decoding settings:
$T\in\{0.5,0.6,0.7,0.8,0.9,1\}$ at $p=1$, and
$p\in\{0.7,0.8,0.9\}$ at $T=0.7$.
For each generation seed, individually normalized MPNet response embeddings
are concatenated across the 100 prompts. One shared Gaussian projection
(seed 42) maps the resulting 76,800-dimensional fingerprint to 128 dimensions,
preserving the projected fingerprint scale.
Each checkpoint--setting cell is fitted with a Gaussian using its sample mean
and an Oracle Approximating Shrinkage covariance estimate.
We report squared Hellinger distance $H^2=1-BC$, where $BC$ is Gaussian
Bhattacharyya affinity. This panel is separate from the primary
30-checkpoint and enlarged 115-checkpoint identification experiments.

\paragraph{Seed banks and comparisons.}
The original figure uses generation seeds 42001--42032 for every model
and setting. We repeat the complete analysis with seeds 43001--43032,
then refit each cell using all 64 samples. The pooled analysis fits each Gaussian to all 64 samples before computing distances.
All models and settings within an analysis share its seed bank and projection.
Temperature contrasts compare $T=1$ with each lower temperature at $p=1$;
top-$p$ contrasts compare $p=1$ with each lower value at $T=0.7$.
Between-checkpoint references use $(T,p)=(0.7,1)$, with 40 within-family
and 780 cross-family pairs. All responses and embeddings are reused
from the archived collections.

\begin{table}[!htbp]\centering\small
\setlength{\tabcolsep}{7pt}
\caption{\textbf{The fitted-distance pattern persists across seed banks and sample sizes.}
Mean squared Hellinger distances for all eight decoding contrasts and the
two checkpoint references. Each decoding row averages 41 checkpoints.}
\label{tab:motivation-seed-checks}
\begin{tabular}{lrrr}\toprule
\textbf{Comparison} & \textbf{Original 32} & \textbf{Independent 32} & \textbf{Pooled 64}\\
\midrule
$T:1\to0.9$, $p=1$ & 0.9094 & 0.8763 & 0.8277 \\
$T:1\to0.8$, $p=1$ & 0.9752 & 0.9695 & 0.9522 \\
$T:1\to0.7$, $p=1$ & 0.9922 & 0.9909 & 0.9852 \\
$T:1\to0.6$, $p=1$ & 0.9958 & 0.9964 & 0.9932 \\
$T:1\to0.5$, $p=1$ & 0.9985 & 0.9986 & 0.9973 \\
\midrule
$p:1\to0.9$, $T=0.7$ & 0.7662 & 0.7425 & 0.6038 \\
$p:1\to0.8$, $T=0.7$ & 0.8552 & 0.8436 & 0.7482 \\
$p:1\to0.7$, $T=0.7$ & 0.9209 & 0.9147 & 0.8608 \\
\midrule
Within-family & 0.9430 & 0.9251 & 0.8886 \\
Cross-family & 0.9881 & 0.9859 & 0.9758 \\
\bottomrule\end{tabular}
\end{table}

\paragraph{Replication results.}
Across all three analyses, $T:1\to0.8$ exceeds the within-family reference,
while $T:1\to0.7$ reaches the cross-family reference scale
(Table~\ref{tab:motivation-seed-checks}). The top-$p$ contrasts remain below
the corresponding pooled within-family reference. These qualitative relations
persist despite changes in the numerical estimates.
Figure~\ref{fig:atlas-motivation} retains the original 32-seed values;
the independent-bank and pooled fits are reported here as sensitivity checks.

\paragraph{Independent-repeat check.}
At $(T,p)=(0.7,1)$, two independent 32-seed banks for the same checkpoint
have mean fitted $H^2=0.728$. This independent-bank comparison reports resampling sensitivity; the main figure retains its original shared-seed distance estimates.

\paragraph{Interpretation and identification.}
The temperature curve compares fitted distributions at each setting with the $T=1$ reference. The $|\Delta T|=0.3$ point represents the $T=1\to0.7$ contrast; SD/IQR summarize checkpoint variation. Section~\ref{sec:robustness} evaluates identification with fixed
enrollment and changed query settings.

\begin{figure}[H]
  \centering
  \includegraphics[width=0.76\linewidth]{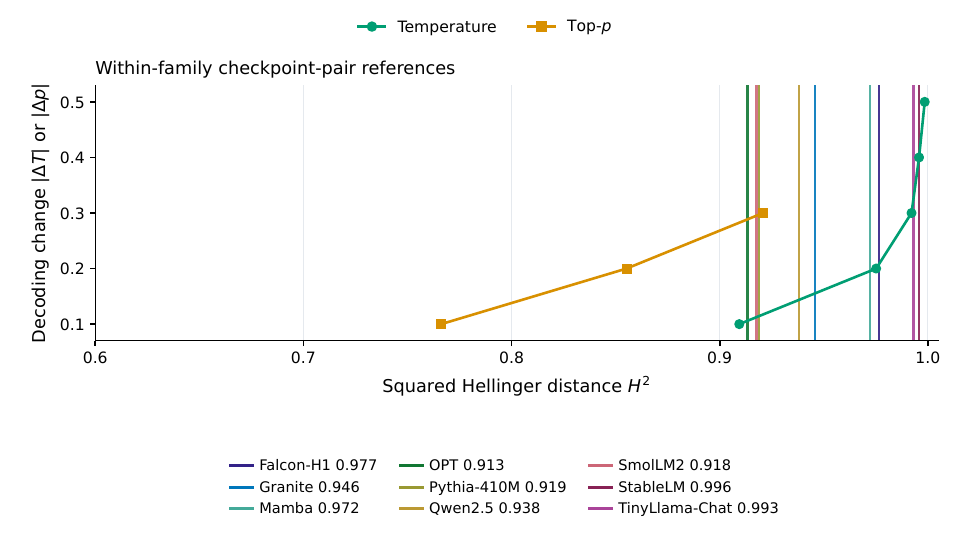}
  \caption{\textbf{Within-family distance references.} Mean original 32-seed
  $H^2$ for checkpoint pairs in each of nine families at $(T,p)=(0.7,1)$;
  curves show the 41-checkpoint decoding contrasts from Figure~\ref{fig:atlas-motivation}(a).}
  \label{fig:atlas-family-refs}
  \vspace{0.5\baselineskip}
  \includegraphics[width=0.86\linewidth]{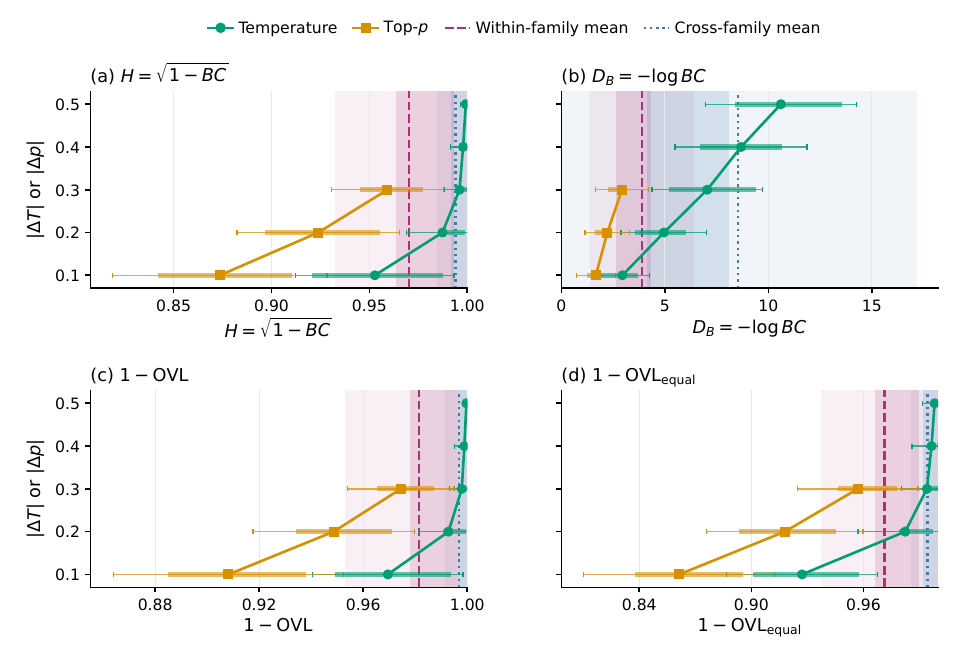}
  \caption{\textbf{Four additional distance metrics.} Original 32-seed fits:
  curves average 41 decoding contrasts per setting. Thin/thick bars show
  SD/IQR; dashed/dotted lines and light/dark bands show pooled within-/cross-family
  means and SD/IQR at $(T,p)=(0.7,1)$.}
  \label{fig:atlas-metric-grid}
\end{figure}

\clearpage
\paragraph{Family-specific reference scales.}
Figure~\ref{fig:atlas-family-refs} retains the within-family reference
panel from the original motivation figure. Each line averages distinct
checkpoint pairs in one family at $(T,p)=(0.7,1)$; the pooled reference
in Figure~\ref{fig:atlas-motivation} averages all within-family pairs.

\begin{figure}[H]
  \centering
  \includegraphics[width=\linewidth,height=0.72\textheight,keepaspectratio]{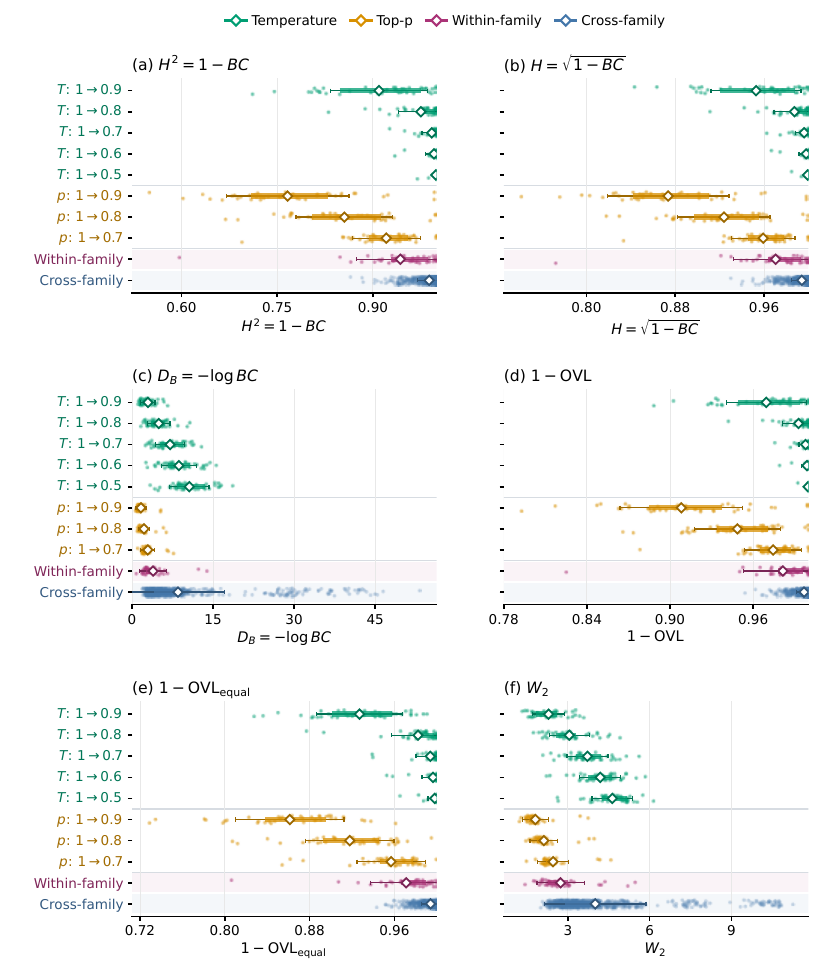}
  \caption{\textbf{Individual comparison distributions for all six metrics.}
  Original 32-seed fits; all 1,148 comparisons per metric are retained:
  eight decoding contrasts across 41 checkpoints, 40 within-family pairs,
  and 780 cross-family pairs. Larger horizontal values indicate greater
  separation. Green/orange denote temperature/top-$p$ contrasts;
  purple/blue denote within-/cross-family pairs. Each dot reports the scalar distance between two fitted distributions.}
  \label{fig:atlas-distribution-grid}
\end{figure}

\paragraph{Additional metric views.}
Figure~\ref{fig:atlas-metric-grid} shows the four metrics omitted from the
main figure. $H=\sqrt{1-BC}$, $D_B=-\log BC$, and $H^2=1-BC$ are dependent transformations of the same fitted affinity. Total variation is
$1-\mathrm{OVL}$, where $\mathrm{OVL}(P,Q)=\int\min\{p(u),q(u)\}\,du$ is
estimated with 4,096 Monte Carlo samples per comparison. The equal-covariance
view replaces both fitted covariances by their mean. Figure~\ref{fig:atlas-distribution-grid}
retains the individual comparisons for all six metrics. All panels use the
original 32-seed fits, separate from the 64-seed refits in
Table~\ref{tab:motivation-seed-checks}.

\paragraph{Individual comparison distributions.}
Each dot in Figure~\ref{fig:atlas-distribution-grid} is one comparison
between two fitted densities. Each decoding row contains 41 values;
the within-family and cross-family rows contain 40 and 780 model pairs.
The 40 within-family pairs come from nine multi-checkpoint families; eleven additional groups contain one checkpoint each. The plot pools pairs by comparison type.
Jitter separates dots visually. Diamonds mark means, thin capped bars
show $\pm$SD, and thick bars show IQR. Pairs sharing checkpoints are
statistically dependent, and the displayed spreads are descriptive.

\clearpage

%% file: sections/appendix-controls.tex
\subsection{Representation, Metric and Storage Controls}\label{app:controls}\label{app:setting-blind}
Table~\ref{tab:representation-controls} compares LLM-DNA at 128, 512, and 1024 coordinates. Entries are means and sample SDs across three projection seeds, combining the two encoder conditions within each seed. The 512-coordinate construction achieves the highest mean Top-1, Top-3, and Macro F1 among the evaluated widths. The primary comparison uses this width with seed 42; Table~\ref{tab:representation-controls} summarizes variation across seeds.

\begin{table}[!htbp]\centering\small
\caption{\textbf{The 512-coordinate LLM-DNA construction leads the evaluated projection widths.} Mean $\pm$ sample SD across three projection seeds; coordinates denote total fingerprint width.}\label{tab:representation-controls}
\input{data/editorial_projection_controls}
\end{table}

\subsection{Discrimination and Descriptive Geometry}\label{app:geometry}
Figure~\ref{fig:main-svm} shows full-space SVM confusion, with IDs in Appendix~\ref{app:visual-ids}. Figure~\ref{fig:main-tsne} gives descriptive projections of the same held-out fingerprints. Both describe the fitted system; classification uses the original representation space.

\begin{figure}[!htbp]\centering
\includegraphics[width=\linewidth]{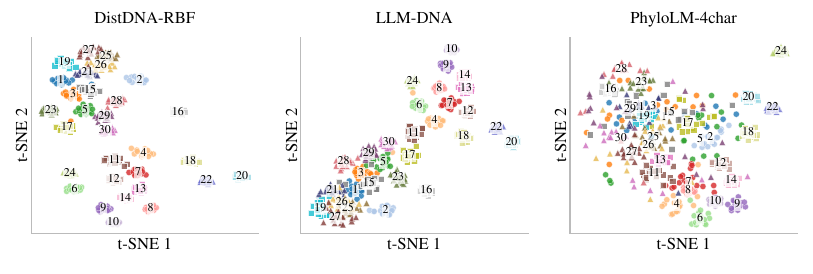}
\caption{\textbf{Checkpoint-specific groups remain visible with MPNet.} Descriptive t-SNE projections of the same 480 held-out queries for \method{}-RBF, LLM-DNA, and PhyloLM-4char; semantic representations use MPNet. Checkpoint coding matches Figure~\ref{fig:main-tsne}.}\label{fig:extended-geometry}
\end{figure}

The t-SNE panels in Figures~\ref{fig:main-tsne} and~\ref{fig:extended-geometry} use $\sqrt{\widehat{\mathrm{MMD}}^2}$, projected-mean Euclidean distance, and $\sqrt{2(1-S_{\mathrm{Nei}})}$ for \method{}-RBF, LLM-DNA, and PhyloLM-4char. Coordinates and projection parameters are frozen; projection sensitivity outputs remain in the artifact archive.

Figure~\ref{fig:extended-geometry} retains representation and secondary-encoder views. All projections use perplexity 30 and seed 20260908, with random initialization, 1,000 iterations and automatic learning rate without principal component analysis (PCA); model IDs are listed in Appendix~\ref{app:visual-ids}.
\paragraph{Random output-space visualization.}
The Random panel projects thirty-dimensional candidate-rank vectors, colored by true checkpoint ID, serving as an output-space diagnostic.

\paragraph{Observation-curve variability.}
At each held-out setting, encoder outcomes and, for enrollment sensitivity, overlapping cyclic-window outcomes are averaged within checkpoint before the eight-group aggregation. Error bars are sample SD across the sixteen resulting setting scores. The fresh allocation collection and primary-cache enrollment collection remain separate.

%% file: data/editorial_projection_controls.tex
\begin{tabular}{lrrr}
\toprule
\textbf{Coordinates}
& \textbf{Top-1 (\%)}
& \textbf{Top-3 (\%)}
& \textbf{Macro F1} \\
\midrule

128
& 83.27 $\pm$ 0.83
& 95.80 $\pm$ 0.40
& 0.8502 $\pm$ 0.0041 \\

512
& \textbf{94.94 $\pm$ 1.27}
& \textbf{99.50 $\pm$ 0.10}
& \textbf{0.9555 $\pm$ 0.0077} \\

1024
& 90.12 $\pm$ 0.80
& 92.56 $\pm$ 0.25
& 0.9169 $\pm$ 0.0051 \\

\bottomrule
\end{tabular}

%% file: sections/appendix-allocation.tex
\subsection{Prompt Breadth and Repetitions}\label{app:allocation}\label{app:fresh-exact30}
Table~\ref{tab:svm-allocation} evaluates the principal enrolled-checkpoint task with eight training and sixteen held-out decoding pairs. Each pair remains fixed within a fingerprint. The first $t$ evaluation prompts and first $R$ draws are selected identically for every method; calibration prompts remain separate. Query cost is $tR$, and enrollment costs $8tR$ responses per checkpoint. Scores average BGE/MPNet and the eight model groups, with encoder-free PhyloLM-4char evaluated once. These exploratory sensitivities use the principal response cache; the primary configuration remains fixed.

\begin{table}[!htbp]\centering\small
\caption{\textbf{Repetition improves SVM identification at matched response budgets.} Primary-cache Top-1 (\%); query cost is $tR$ and enrollment cost is $8tR$ responses/model.}\label{tab:svm-allocation}
\input{data/svm_allocation}
\end{table}
At both reduced query budgets, allocating more draws to fewer prompts gives higher point estimates for each method in this nested panel. The finding characterizes the evaluated allocations and prompt panel.

\paragraph{Statistical confidence on accuracy.}
Table~\ref{tab:svm-confidence} varies prompt/draw subsets over five fixed seeds, averaging them and both encoders before uncertainty estimation. A crossed bootstrap resamples the sixteen test settings and checkpoints within each fixed group. A common maximum-deviation margin supplies approximate simultaneous 95\% lower bounds over the original 27 method/allocation cells. This quantifies uncertainty in aggregate accuracy, and it is conditional on the enrolled roster, prompt pool and fitted enrollment data.

\begin{table}[!htbp]\centering\small
\caption{\textbf{Conditional lower bounds quantify the response budget needed for target accuracy.} Top-1 followed by its simultaneous 95\% lower bound in parentheses (\%), with prompt/draw subset averaging. Enrollment costs eight times the listed query calls.}\label{tab:svm-confidence}
\input{data/svm_confidence}
\end{table}

Within this tested grid, each retained \method{} construction first meets an 80\% accuracy target at 40 query calls and a 90\% target at 80 calls under the simultaneous lower bound. The simultaneous lower bounds remain below 95\% for every tested allocation. These are finite-grid, exploratory budget estimates.
The completed fresh-draw study and its distinct simultaneous family are reported in Appendix~\ref{app:fresh-completion}.

%% file: data/svm_allocation.tex
\begin{tabular}{lrrrrr}
\toprule
$t\times R$ & \textbf{Calls} & \textbf{RBF} & \textbf{Linear} & \textbf{LLM-DNA} & \textbf{PhyloLM-4char} \\
\midrule
$5\times 4$ & 20 & 83.43 & 84.18 & 77.38 & 58.59 \\
$10\times 2$ & 20 & 77.47 & 77.12 & 67.82 & 52.96 \\
$20\times 1$ & 20 & 73.73 & 72.38 & 56.58 & 49.35 \\
$10\times 4$ & 40 & 94.34 & 94.61 & 89.63 & 69.56 \\
$20\times 2$ & 40 & 92.42 & 92.45 & 79.65 & 68.59 \\
$20\times 4$ & 80 & 98.96 & 98.96 & 95.31 & 86.62 \\
\bottomrule
\end{tabular}

%% file: data/svm_confidence.tex
\begin{tabular}{lrrr}
\toprule
\textbf{$t\times R$} & \textbf{Calls} & \textbf{RBF} & \textbf{Linear} \\
\midrule
$5\times 1$ & 5 & 41.98 (34.15) & 41.02 (33.19) \\
$5\times 2$ & 10 & 66.73 (58.90) & 66.29 (58.46) \\
$5\times 4$ & 20 & 85.17 (77.34) & 84.97 (77.14) \\
$10\times 1$ & 10 & 57.03 (49.20) & 55.92 (48.09) \\
$10\times 2$ & 20 & 80.99 (73.15) & 80.61 (72.77) \\
$10\times 4$ & 40 & 95.36 (87.53) & 95.41 (87.58) \\
$20\times 1$ & 20 & 73.35 (65.52) & 72.33 (64.50) \\
$20\times 2$ & 40 & 93.30 (85.47) & 93.25 (85.42) \\
$20\times 4$ & 80 & 98.96 (91.13) & 98.96 (91.13) \\
\bottomrule
\end{tabular}

%% file: sections/appendix-parameters.tex
\subsection{SVM Parameters and RBF Bandwidth}\label{app:parameters}
\begin{figure}[!htbp]\centering
\includegraphics[width=\linewidth]{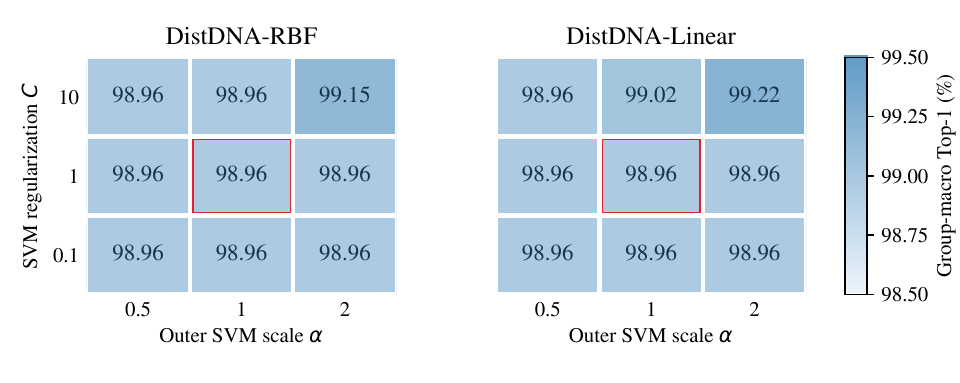}
\caption{\textbf{\method{} is insensitive to the tested SVM regularization and scale.} Top-1 (\%), averaged over BGE and MPNet, with 80 query responses and 640 enrollment responses/model. Panels share a 98.5--99.5\% accuracy scale; red outlines mark the unchanged primary setting $C=\alpha=1$.}\label{fig:svm-parameters}
\end{figure}

\method{} is insensitive to the SVM regularization $C$ and outer-kernel scale $\alpha$ over the evaluated grid (Figure~\ref{fig:svm-parameters}). Across $C\in\{0.1,1,10\}$ and $\alpha\in\{0.5,1,2\}$, Top-1 varies by only 0.20 percentage points for \method{}-RBF and 0.26 for \method{}-Linear. This stability supports the fixed primary choice $C=\alpha=1$ across the tested ranges. The outer kernel is $\exp[-d^2/(\alpha s)]$, where $s$ is the median positive training squared distance. All cells share the same response banks, with 80 query responses and 640 enrollment responses per model, averaged over BGE and MPNet.

%% file: sections/appendix-variance.tex
\subsection{Finite-Sample Distance Correction}\label{app:variance-control}
For independent arms, Theorem~\ref{thm:repetition} decomposes expected empirical squared distance into population squared distance plus query and candidate variance divided by their repetition counts. For a fixed query, the query-variance term cancels in pairwise differences of expected candidate scores, while unequal candidate variances contribute a ranking offset. Subtracting estimated candidate variance diagnoses ranking bias while potentially increasing estimator variance and producing negative corrected scores. The exact-kernel counterpart is the MMD U-statistic (up to the query-constant term).

\begin{figure}[!htb]\centering
\includegraphics[width=\linewidth]{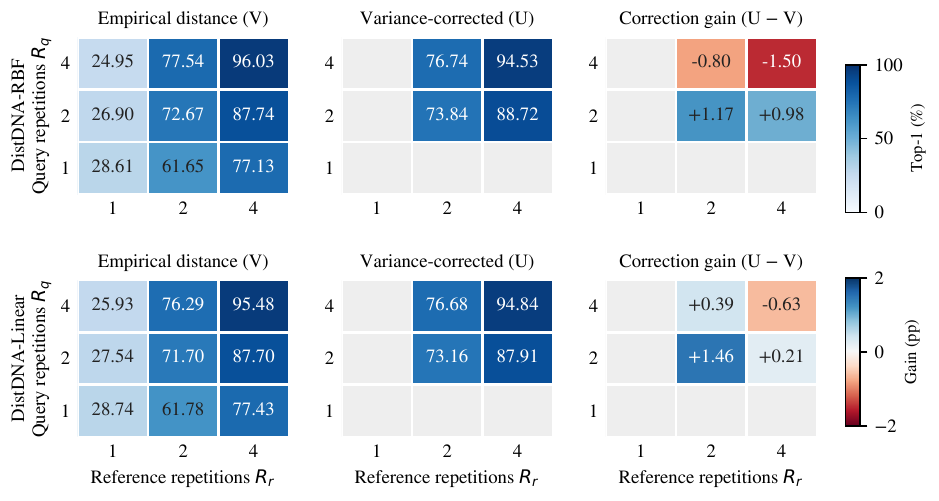}
\caption{\textbf{Variance correction has mixed effects on nearest-reference identification.} Top-1 for empirical distances (V) and corrected distances (U), with correction gains in percentage points. Each fingerprint uses 20 prompts; queries cost $20R_q$ responses and enrollment costs $160R_r$ per model. Gray cells mark corrections requiring at least two draws; identification uses nearest-reference matching. }\label{fig:nearest-variance}
\end{figure}

Figure~\ref{fig:nearest-variance} uses twenty prompts and independently varies query and reference repetition counts. Identification uses the nearest of eight individual enrolled reference fingerprints per checkpoint. Enrollment costs $8\times20R_r$ responses per checkpoint and queries cost $20R_q$. The U score subtracts both empirical sample-mean variance terms and is evaluated only when both repetition counts are at least two. Negative scores are retained. Averaging uses BGE/MPNet and the eight reporting groups. Corrections yield both gains and losses; the V-statistic remains the primary estimator.

%% file: sections/appendix-fresh.tex
\subsection{Fresh Draws, Allocation, Domain and Interface}\label{app:fresh-completion}
The completed follow-up contains 261,200 response records and 162 fitted SVM evaluations, all retained. The original 30 checkpoints and eight enrollment/sixteen test decoding pairs are unchanged. Dolly generation uses a new seed namespace over 80 evaluation prompts and 20 calibration prompts; the first 20 evaluation prompts overlap the primary study. Tiered repetition counts support five declared 80-response allocations and ten lower-budget cells. Enrollment costs eight times query cost. Generation uses float32, plain-text prompts, the original response boundary and one attempt; all 182 empty outputs are retained. Encoders, seed-42 LLM-DNA, four-character PhyloLM-4char, and SVM policy follow the primary constructions. Each condition is separately enrolled and calibrated on its admissible calibration prompts.

All 49 source/manifest hashes in the collection lock were verified, as were response and embedding hashes, disjoint calibration/evaluation indices, output counts and pinned revisions. Reconstructing every per-fit aggregate from 72,864 saved prediction rows reproduces all 162 fit results and 90 encoder-aggregated results.

\begin{table}[!htb]\centering\small
\caption{\textbf{The fresh collection reproduces the benefit of repeated responses.} Complete SVM allocation grid, with Top-1 (\%). Calls are per query; enrollment costs eight times this amount per checkpoint.}\label{tab:fresh-full}
\input{data/fresh_full_allocation}
\end{table}
Table~\ref{tab:fresh-full} reports the complete declared allocation grid, including lower-budget cells.

\paragraph{Conditional accuracy bounds.}
The frozen fresh-study procedure averages encoder outcomes before 5,000 crossed setting/checkpoint-within-group bootstrap replicates (seed 2026091817). Its simultaneous one-sided basic-bootstrap lower bounds cover the original 75 method/allocation cells, including external baselines and the archived random-feature approximation; all bounds retain this original comparison family.
The common upper-error margin is \FreshConfidenceMargin{} percentage points. Table~\ref{tab:fresh-confidence} reports every lower bound.
For each retained \method{} construction, the smallest tested qualifying budget at an 80\% target is 40 query calls (320 enrollment/model). These bounds quantify aggregate accuracy conditional on the observed roster, prompt pool, fitted enrollment, and tested allocation grid.
\begin{table}[!htbp]\centering\small
\caption{\textbf{The fresh-study bounds support an 80\% target at a 40-response query budget.} Simultaneous 95\% lower bounds (\%) use the full 75-cell family; point estimates appear in Table~\ref{tab:fresh-full}.}\label{tab:fresh-confidence}
\input{data/fresh_confidence}
\end{table}

\paragraph{Domain and interface conditions.}

The authored procedural domain uses separate calibration/evaluation prompts and its own labeled enrollment. The plain/native-chat comparison uses precisely the same thirteen eligible checkpoints, prompts and decoding pairs; the two arms share generation seeds as a paired intervention. Eligibility requires a native template that accepts the fixed system/user messages.
Interface effects include system-message wrapping.
Table~\ref{tab:fresh-domains} includes all methods and both interface arms; the small eligible gallery is near saturation.
The numerical results are reported in Table~\ref{tab:fresh-domains}; this subsection specifies the collection protocol.
\begin{table}[!htbp]\centering\small
\caption{\textbf{Identification remains accurate after separate enrollment within each condition.} Top-1 (\%); procedural and random-CFG use thirty checkpoints and paired plain/chat uses thirteen. Each uses 80 query responses and 640 enrollment responses/model.}\label{tab:fresh-domains}
\input{data/fresh_domains}
\end{table}

\subsection{Fixed-Classifier Decoding Evaluation}
\label{app:stress-plan}

The completed query-only evaluation retains the original
30-model enrollment bank, calibration, encoders and fitted
classifiers. The stochastic grid crosses
$T\in\{0.71,0.85,1.0\}$ with
$p\in\{0.8,0.9,1.0\}$.
Each condition uses twenty prompts with four responses
per prompt, totaling 21,600 stochastic responses.
A separate greedy condition adds 600 responses, with one
response per prompt; each deterministic output is retained once.

%% file: data/fresh_full_allocation.tex
\begin{tabular}{lrrrrr}
\toprule
\textbf{$t\times R$} & \textbf{Calls} & \textbf{Linear} & \textbf{RBF} & \textbf{LLM-DNA} & \textbf{PhyloLM-4char} \\
\midrule
$5\times1$ & 5 & 38.72 & 37.96 & 35.90 & 22.33 \\
$5\times2$ & 10 & 62.34 & 63.74 & 57.06 & 41.83 \\
$10\times1$ & 10 & 54.07 & 54.00 & 47.15 & 34.70 \\
$5\times4$ & 20 & 82.54 & 82.45 & 79.74 & 57.65 \\
$10\times2$ & 20 & 78.76 & 78.61 & 71.44 & 50.85 \\
$20\times1$ & 20 & 73.18 & 74.25 & 60.30 & 43.10 \\
$5\times8$ & 40 & 95.31 & 95.39 & 93.73 & 77.54 \\
$10\times4$ & 40 & 92.90 & 93.46 & 88.75 & 70.48 \\
$20\times2$ & 40 & 92.02 & 92.59 & 82.29 & 66.08 \\
$40\times1$ & 40 & 84.67 & 85.51 & 63.70 & 54.69 \\
$5\times16$ & 80 & 99.77 & 99.77 & 99.45 & 90.95 \\
$10\times8$ & 80 & 98.70 & 98.57 & 97.62 & 90.76 \\
$20\times4$ & 80 & 98.80 & 98.99 & 95.41 & 86.52 \\
$40\times2$ & 80 & 97.01 & 97.46 & 85.04 & 81.18 \\
$80\times1$ & 80 & 92.20 & 92.37 & 64.53 & 65.23 \\
\bottomrule
\end{tabular}

%% file: data/fresh_confidence.tex
\begin{tabular}{lrrrrr}
\toprule
\textbf{$t\times R$} & \textbf{Calls} & \textbf{Linear} & \textbf{RBF} & \textbf{LLM-DNA} & \textbf{PhyloLM-4char} \\
\midrule
$5\times1$ & 5 & 26.81 & 26.04 & 23.99 & 10.42 \\
$5\times2$ & 10 & 50.42 & 51.82 & 45.15 & 29.92 \\
$10\times1$ & 10 & 42.15 & 42.09 & 35.24 & 22.79 \\
$5\times4$ & 20 & 70.62 & 70.54 & 67.82 & 45.74 \\
$10\times2$ & 20 & 66.85 & 66.70 & 59.52 & 38.93 \\
$20\times1$ & 20 & 61.26 & 62.34 & 48.39 & 31.18 \\
$5\times8$ & 40 & 83.40 & 83.48 & 81.82 & 65.62 \\
$10\times4$ & 40 & 80.99 & 81.54 & 76.84 & 58.56 \\
$20\times2$ & 40 & 80.11 & 80.68 & 70.38 & 54.17 \\
$40\times1$ & 40 & 72.75 & 73.60 & 51.79 & 42.77 \\
$5\times16$ & 80 & 87.86 & 87.86 & 87.53 & 79.04 \\
$10\times8$ & 80 & 86.78 & 86.65 & 85.71 & 78.84 \\
$20\times4$ & 80 & 86.88 & 87.08 & 83.50 & 74.61 \\
$40\times2$ & 80 & 85.09 & 85.55 & 73.13 & 69.27 \\
$80\times1$ & 80 & 80.29 & 80.45 & 52.62 & 53.32 \\
\bottomrule
\end{tabular}

%% file: data/fresh_domains.tex
\begin{tabular}{lrrrr}
\toprule
\textbf{Method} & \textbf{Procedural (30)} & \textbf{Random CFG (30)}
& \textbf{Plain (13)} & \textbf{Chat (13)} \\
\midrule
\distdnalinear{} & 96.44 & 89.52 & 99.74 & 100.00 \\
\distdnaexact{}   & 95.88 & 87.91 & 99.74 & 100.00 \\
LLM-DNA            & 92.30 & 77.29 & 99.80 & 100.00 \\
PhyloLM-4char      & 90.30 & 64.75 & 100.00 & 100.00 \\
\bottomrule
\end{tabular}

%% file: sections/limitations.tex
Our evaluations identify enrolled checkpoint entries under specified prompt, decoding, and collection protocols. The original thirty-checkpoint experiment weights its eight reporting groups equally, while the enlarged 115-entry gallery uses equal-entry weighting. The evaluated entries are pinned repository revisions; weight-equivalent aliases require a separate audit. \method{}-Linear matches RBF Top-1 accuracy in the original thirty-checkpoint supervised study. In the main 115-entry grid, RBF has higher Top-1 under both readouts, while Linear has higher supervised Top-3. The comparisons match response budgets and readout policies, while fingerprint storage and native baseline operating regimes differ.

The original thirty-checkpoint collection uses short plain-text generations. Domain and native-chat experiments enroll and fit classifiers separately within each condition. The fixed-classifier grid covers the reported temperature/top-$p$ cells and a separate greedy endpoint. These experiments characterize the stated conditions; open-set rejection, zero-shot interface transfer, and arbitrary deployment settings require separate evaluation. Response encoders can discard textual distinctions. Statistical intervals condition on the observed roster, prompt pool, and fitted enrollment, and the finite allocation grids characterize the evaluated designs. The theory establishes representation and sampling properties for checkpoint--decoding pairs; identification performance is assessed empirically.

%% file: sections/broader-impact.tex
Behavioral fingerprints can support model inventory, API verification, regression monitoring, and provenance investigations without access to weights. In a positive deployment, an operator could enroll candidate checkpoints and detect an accidental serving change using only ordinary text queries.

The same capability can reveal a provider's model choices without consent or be overinterpreted as evidence of misconduct. Claims about copied weights, common training data, license violations, or shared ancestry require evidence beyond response similarity.
False attribution can cause reputational and legal harm, especially when a candidate checkpoint is absent from the gallery. Consequential use therefore requires corroborating documentation or parameter-level evidence, calibrated uncertainty, and a process for contesting an attribution. Query sets and stored fingerprints should also be protected where they expose proprietary deployment information.

%% file: sections/appendix-inventory.tex
\subsection{Models and Revisions}
Table~\ref{tab:model-inventory} lists the 115 repository/checkpoint entries used in the enlarged gallery. Rows are sorted by repository owner and numbered in display order. The inventory counts pinned repository entries.
\input{data/editorial_model_inventory}
\subsection{Primary Figure ID Key}\label{app:visual-ids}
The following original IDs identify checkpoints in the t-SNE and confusion matrices.
\begingroup\small
\input{data/primary_figure_ids}
\endgroup

%% file: data/editorial_model_inventory.tex
\begingroup
\small
\setlength{\LTcapwidth}{\linewidth}
\makeatletter\g@addto@macro\UrlBreaks{\do\a\do\b\do\c\do\d\do\e\do\f\do\g\do\h\do\i\do\j\do\k\do\l\do\m\do\n\do\o\do\p\do\q\do\r\do\s\do\t\do\u\do\v\do\w\do\x\do\y\do\z\do\A\do\B\do\C\do\D\do\E\do\F\do\G\do\H\do\I\do\J\do\K\do\L\do\M\do\N\do\O\do\P\do\Q\do\R\do\S\do\T\do\U\do\V\do\W\do\X\do\Y\do\Z}\makeatother
\setlength{\tabcolsep}{3pt}
\begin{longtable}{r p{0.20\linewidth} p{0.49\linewidth} l}
\caption{\textbf{The 115-model gallery.}}\label{tab:model-inventory}\\
\toprule \textbf{ID} & \textbf{Company / org.} & \textbf{Checkpoint} & \textbf{Revision prefix} \\\midrule\endfirsthead
\toprule \textbf{ID} & \textbf{Company / org.} & \textbf{Checkpoint} & \textbf{Revision prefix} \\\midrule\endhead
\bottomrule\endlastfoot
1 & \nolinkurl{ahxt} & \nolinkurl{LiteLlama-460M-1T} & \texttt{77b8a976440e} \\
2 & \nolinkurl{akhauriyash} & \nolinkurl{DeepSeek-R1-Distill-Qwen-1.5B-SpeculativeReasoner} & \texttt{130021037bab} \\
3 & \nolinkurl{Azurro} & \nolinkurl{APT3-1B-Base} & \texttt{a9dbe5e7cc05} \\
4 & \nolinkurl{bigscience} & \nolinkurl{bloom-1b1} & \texttt{eb3dd7399312} \\
5 & \nolinkurl{bigscience} & \nolinkurl{bloom-560m} & \texttt{ac2ae5fab2ce} \\
6 & \nolinkurl{bigscience} & \nolinkurl{bloomz-1b1} & \texttt{3c4e8ecc1fbe} \\
7 & \nolinkurl{bigscience} & \nolinkurl{bloomz-560m} & \texttt{a2845d7e13dd} \\
8 & \nolinkurl{bigscience} & \nolinkurl{bloomz-7b1} & \texttt{fa72440b228d} \\
9 & \nolinkurl{CEIA-POSITIVO2} & \nolinkurl{Qwen-1.7B-pt-capado} & \texttt{2e4a83201a7b} \\
10 & \nolinkurl{DeepMount00} & \nolinkurl{Alireo-400m-instruct-v0.1} & \texttt{4457428d3201} \\
11 & \nolinkurl{e-palmisano} & \nolinkurl{Qwen2-0.5B-ITA-IT} & \texttt{2a4565f5d9a6} \\
12 & \nolinkurl{Edcastro} & \nolinkurl{DeepSeek-R1-Distill-Qwen-1.5B-edcastr_JavaScript-v8} & \texttt{6d9bb37f736a} \\
13 & \nolinkurl{EleutherAI} & \nolinkurl{pythia-160m-seed1} & \texttt{37cfeed3312c} \\
14 & \nolinkurl{EleutherAI} & \nolinkurl{pythia-160m-seed2} & \texttt{119737fa89f0} \\
15 & \nolinkurl{EleutherAI} & \nolinkurl{pythia-160m-seed3} & \texttt{1b49c41a05c5} \\
16 & \nolinkurl{EleutherAI} & \nolinkurl{pythia-160m-seed4} & \texttt{352dff55fd10} \\
17 & \nolinkurl{EleutherAI} & \nolinkurl{pythia-160m-seed5} & \texttt{136e28f57548} \\
18 & \nolinkurl{EleutherAI} & \nolinkurl{pythia-160m-seed6} & \texttt{78e11a97c628} \\
19 & \nolinkurl{EleutherAI} & \nolinkurl{pythia-160m-seed8} & \texttt{2149b1efc141} \\
20 & \nolinkurl{EleutherAI} & \nolinkurl{pythia-160m-seed9} & \texttt{08020a8fabd0} \\
21 & \nolinkurl{EleutherAI} & \nolinkurl{pythia-410m-seed1} & \texttt{33803c4f5a1e} \\
22 & \nolinkurl{EleutherAI} & \nolinkurl{pythia-410m-seed2} & \texttt{da2549c8d3e0} \\
23 & \nolinkurl{EleutherAI} & \nolinkurl{pythia-410m-seed3} & \texttt{1dd84c3d638f} \\
24 & \nolinkurl{EleutherAI} & \nolinkurl{pythia-410m-seed4} & \texttt{57c8ecb67375} \\
25 & \nolinkurl{EleutherAI} & \nolinkurl{pythia-410m-seed5} & \texttt{8f2c179eee40} \\
26 & \nolinkurl{EleutherAI} & \nolinkurl{pythia-410m-seed6} & \texttt{6a2463662069} \\
27 & \nolinkurl{EleutherAI} & \nolinkurl{pythia-410m-seed7} & \texttt{84f51f7862c3} \\
28 & \nolinkurl{EleutherAI} & \nolinkurl{pythia-410m-seed8} & \texttt{368c5c53fda3} \\
29 & \nolinkurl{EleutherAI} & \nolinkurl{pythia-410m-seed9} & \texttt{509afc6f4768} \\
30 & \nolinkurl{FabienRoger} & \nolinkurl{cot_5k} & \texttt{1feab8ac771d} \\
31 & \nolinkurl{facebook} & \nolinkurl{opt-1.3b} & \texttt{3f5c25d0bc63} \\
32 & \nolinkurl{facebook} & \nolinkurl{opt-125m} & \texttt{27dcfa74d334} \\
33 & \nolinkurl{facebook} & \nolinkurl{opt-350m} & \texttt{08ab08cc4b72} \\
34 & \nolinkurl{facebook} & \nolinkurl{opt-iml-1.3b} & \texttt{2217fd558c1f} \\
35 & \nolinkurl{GritLM} & \nolinkurl{GritLM-7B} & \texttt{cb7f7ffb99c0} \\
36 & \nolinkurl{h2oai} & \nolinkurl{h2o-danube3-4b-base} & \texttt{6bdf2f1e3171} \\
37 & \nolinkurl{HuggingFaceH4} & \nolinkurl{mistral-7b-sft-beta} & \texttt{d7b9d226fb91} \\
38 & \nolinkurl{HuggingFaceTB} & \nolinkurl{SmolLM2-1.7B} & \texttt{effd688a1292} \\
39 & \nolinkurl{HuggingFaceTB} & \nolinkurl{SmolLM2-1.7B-IT} & \texttt{31b70e2e869a} \\
40 & \nolinkurl{HuggingFaceTB} & \nolinkurl{SmolLM2-135M} & \texttt{93efa2f097d5} \\
41 & \nolinkurl{HuggingFaceTB} & \nolinkurl{SmolLM2-135M-IT} & \texttt{12fd25f77366} \\
42 & \nolinkurl{HuggingFaceTB} & \nolinkurl{SmolLM2-360M} & \texttt{f8027fd0eaee} \\
43 & \nolinkurl{HuggingFaceTB} & \nolinkurl{SmolLM2-360M-IT} & \texttt{a10cc1512eab} \\
44 & \nolinkurl{Ihor} & \nolinkurl{Text2Graph-R1-Qwen2.5-0.5b} & \texttt{a1727ba45774} \\
45 & \nolinkurl{Jiayi-Pan} & \nolinkurl{Tiny-Vicuna-1B} & \texttt{94f739d96610} \\
46 & \nolinkurl{katanemo} & \nolinkurl{Arch-Router-1.5B} & \texttt{0490c1af1085} \\
47 & \nolinkurl{KBlueLeaf} & \nolinkurl{TIPO-200M-ft2} & \texttt{1ff5c809bd44} \\
48 & \nolinkurl{lamini} & \nolinkurl{lamini_docs_finetuned} & \texttt{7fb0b3eb4373} \\
49 & \nolinkurl{llm-jp} & \nolinkurl{llm-jp-3-150m} & \texttt{b112feef602f} \\
50 & \nolinkurl{lmsys} & \nolinkurl{longchat-7b-v1.5-32k} & \texttt{389eea141540} \\
51 & \nolinkurl{LorenaYannnnn} & \nolinkurl{20260217-Qwen3-0.6B_grpo_warmup_16000_episodes_seed_42} & \texttt{4ffe25da00d7} \\
52 & \nolinkurl{LorenaYannnnn} & \nolinkurl{longer_response-Qwen3-0.6B-baseline_all_tokens-seed_0} & \texttt{fbdbb884635f} \\
53 & \nolinkurl{microsoft} & \nolinkurl{phi-1} & \texttt{d4c0adcb065e} \\
54 & \nolinkurl{microsoft} & \nolinkurl{phi-1_5} & \texttt{77aa61eeac94} \\
55 & \nolinkurl{microsoft} & \nolinkurl{phi-2} & \texttt{810d367871c1} \\
56 & \nolinkurl{migtissera} & \nolinkurl{Tess-XS-v1-3-yarn-128K} & \texttt{72d393d13f1b} \\
57 & \nolinkurl{mii-llm} & \nolinkurl{maestrale-chat-v0.4-beta} & \texttt{3a0e5e58f3e7} \\
58 & \nolinkurl{mii-llm} & \nolinkurl{minerva-chat-v0.1-alpha-sft} & \texttt{840db0afa3b7} \\
59 & \nolinkurl{Nanbeige} & \nolinkurl{Nanbeige4-3B-Base} & \texttt{8eeb73458b8e} \\
60 & \nolinkurl{NbAiLab} & \nolinkurl{nb-notram-llama-3.2-1b-instruct} & \texttt{566f459c355a} \\
61 & \nolinkurl{nickypro} & \nolinkurl{tinyllama-110M-fp32} & \texttt{dec0099ae654} \\
62 & \nolinkurl{numind} & \nolinkurl{NuExtract-1.5-tiny} & \texttt{63e2e80c804d} \\
63 & \nolinkurl{nvidia} & \nolinkurl{AceInstruct-1.5B} & \texttt{1e3d02075fcf} \\
64 & \nolinkurl{nvidia} & \nolinkurl{AceMath-1.5B-IT} & \texttt{166818c371ea} \\
65 & \nolinkurl{OPI-PG} & \nolinkurl{Qra-1b} & \texttt{a273489ac073} \\
66 & \nolinkurl{p1atdev} & \nolinkurl{dart-v1-sft} & \texttt{dd5a3f34f3ba} \\
67 & \nolinkurl{princeton-nlp} & \nolinkurl{Sheared-LLaMA-2.7B} & \texttt{2f157a0306b7} \\
68 & \nolinkurl{prithivMLmods} & \nolinkurl{Llama-SmolTalk-3.2-1B-IT} & \texttt{160637645e65} \\
69 & \nolinkurl{Qwen} & \nolinkurl{Qwen1.5-7B-Chat} & \texttt{5f4f5e69ac7f} \\
70 & \nolinkurl{Qwen} & \nolinkurl{Qwen2.5-0.5B} & \texttt{060db6499f32} \\
71 & \nolinkurl{Qwen} & \nolinkurl{Qwen2.5-0.5B-IT} & \texttt{7ae557604adf} \\
72 & \nolinkurl{Qwen} & \nolinkurl{Qwen2.5-1.5B} & \texttt{8faed761d45a} \\
73 & \nolinkurl{Qwen} & \nolinkurl{Qwen2.5-1.5B-IT} & \texttt{989aa7980e4c} \\
74 & \nolinkurl{Qwen} & \nolinkurl{Qwen2.5-Coder-0.5B} & \texttt{8123ea2e9354} \\
75 & \nolinkurl{Qwen} & \nolinkurl{Qwen2.5-Coder-0.5B-IT} & \texttt{ea3f2471cf1b} \\
76 & \nolinkurl{Qwen} & \nolinkurl{Qwen2.5-Coder-1.5B} & \texttt{df3ce67c0e24} \\
77 & \nolinkurl{Qwen} & \nolinkurl{Qwen2.5-Coder-1.5B-IT} & \texttt{2e1fd397ee46} \\
78 & \nolinkurl{Qwen} & \nolinkurl{Qwen3-0.6B} & \texttt{c1899de289a0} \\
79 & \nolinkurl{Qwen} & \nolinkurl{Qwen3-1.7B} & \texttt{70d244cc86cc} \\
80 & \nolinkurl{Qwen} & \nolinkurl{Qwen3-4B} & \texttt{1cfa9a720891} \\
81 & \nolinkurl{raicrits} & \nolinkurl{Hermes7b_ITA} & \texttt{d0547db358a1} \\
82 & \nolinkurl{ReDiX} & \nolinkurl{Qwen2.5-0.5B-IT-ITA} & \texttt{7da45e45ee3f} \\
83 & \nolinkurl{Salesforce} & \nolinkurl{xLAM-1b-fc-r} & \texttt{20e8e2d0d15a} \\
84 & \nolinkurl{sampluralis} & \nolinkurl{llama-sft-sgd} & \texttt{f312c8951af5} \\
85 & \nolinkurl{SeacomSrl} & \nolinkurl{SeaQwen2-0.5B} & \texttt{cc4a02b70473} \\
86 & \nolinkurl{seeweb} & \nolinkurl{SeewebLLM-it} & \texttt{8e69389cbe89} \\
87 & \nolinkurl{stabilityai} & \nolinkurl{stable-code-3b} & \texttt{5cee3fa2905e} \\
88 & \nolinkurl{stabilityai} & \nolinkurl{StableLM-1.6B base} & \texttt{f499ead74c53} \\
89 & \nolinkurl{stabilityai} & \nolinkurl{StableLM-1.6B base-chat} & \texttt{f3fe67057c27} \\
90 & \nolinkurl{stabilityai} & \nolinkurl{StableLM-1.6B chat} & \texttt{2f275b1127d5} \\
91 & \nolinkurl{stas122} & \nolinkurl{Stentor-Big} & \texttt{214407c2b629} \\
92 & \nolinkurl{suayptalha} & \nolinkurl{Qwen3-0.6B-Code-Expert} & \texttt{02c021dfc15b} \\
93 & \nolinkurl{tiiuae} & \nolinkurl{Falcon-H1-0.5B-Base} & \texttt{59fb76e8c5d3} \\
94 & \nolinkurl{tiiuae} & \nolinkurl{Falcon-H1-0.5B-IT} & \texttt{8f2587ca06bf} \\
95 & \nolinkurl{tiiuae} & \nolinkurl{Falcon-H1-1.5B-Base} & \texttt{06d733026625} \\
96 & \nolinkurl{tiiuae} & \nolinkurl{Falcon-H1-1.5B-Deep-IT} & \texttt{b6648636ddc9} \\
97 & \nolinkurl{tiiuae} & \nolinkurl{Falcon-H1-1.5B-IT} & \texttt{80ebc50d7799} \\
98 & \nolinkurl{tiiuae} & \nolinkurl{Falcon-H1-3B-Base} & \texttt{c096902c69be} \\
99 & \nolinkurl{tiiuae} & \nolinkurl{Falcon-H1-7B-IT} & \texttt{41e72f27effb} \\
100 & \nolinkurl{tiiuae} & \nolinkurl{Falcon-H1R-7B} & \texttt{a6f74bf18138} \\
101 & \nolinkurl{TinyLlama} & \nolinkurl{TinyLlama-1.1B base} & \texttt{59f6f375b26b} \\
102 & \nolinkurl{TinyLlama} & \nolinkurl{TinyLlama-1.1B chat} & \texttt{fe8a4ea1ffed} \\
103 & \nolinkurl{TinyLlama} & \nolinkurl{TinyLlama-1.1B-Chat-v0.1} & \texttt{7abc14e7779e} \\
104 & \nolinkurl{TinyLlama} & \nolinkurl{TinyLlama-1.1B-Chat-v0.4} & \texttt{5378798738b1} \\
105 & \nolinkurl{TinyLlama} & \nolinkurl{TinyLlama-1.1B-Chat-v0.6} & \texttt{bf9ae1c8bf02} \\
106 & \nolinkurl{togethercomputer} & \nolinkurl{Llama-2-7B-32K-IT} & \texttt{d27380af0032} \\
107 & \nolinkurl{unsloth} & \nolinkurl{DeepSeek-R1-Distill-Qwen-1.5B} & \texttt{81ae06f3626e} \\
108 & \nolinkurl{unsloth} & \nolinkurl{gemma-2-2b-it} & \texttt{457f2e15bf55} \\
109 & \nolinkurl{unsloth} & \nolinkurl{Qwen2-1.5B-IT} & \texttt{bb43ba0d1578} \\
110 & \nolinkurl{unsloth} & \nolinkurl{Qwen2.5-3B-IT} & \texttt{7548fff1f997} \\
111 & \nolinkurl{unsloth} & \nolinkurl{Qwen3-0.6B-Base} & \texttt{bb9061da4d54} \\
112 & \nolinkurl{unsloth} & \nolinkurl{SmolLM-360M} & \texttt{edd5803235f0} \\
113 & \nolinkurl{utter-project} & \nolinkurl{EuroLLM-1.7B-IT} & \texttt{a25c7fa65fc2} \\
114 & \nolinkurl{Writer} & \nolinkurl{palmyra-mini-thinking-a} & \texttt{64444a2e8a36} \\
115 & \nolinkurl{znhoughton} & \nolinkurl{opt-c4-350m-seed964} & \texttt{ffc99538d210} \\
\end{longtable}
\endgroup

\paragraph{Native-chat service expansion inventory}
The native-mode family expansion in Appendix~F.13 evaluates 86 live production API service IDs across six model families:

\begin{center}
\footnotesize
\renewcommand{\arraystretch}{1.25}
\begin{tabular}{@{}l >{\raggedright\arraybackslash}p{0.79\textwidth}@{}}
\toprule
\textbf{Family (Count)} & \textbf{Evaluated API Service IDs} \\
\midrule
DeepSeek (13) &
\texttt{deepseek/deepseek-chat},
\texttt{deepseek/deepseek-chat-v3-0324},
\texttt{deepseek/deepseek-chat-v3.1},
\texttt{deepseek/deepseek-r1-0528},
\texttt{deepseek/deepseek-v3.2},
\texttt{deepseek/deepseek-v3.2-exp},
\texttt{deepseek/deepseek-v4-flash},
\texttt{deepseek/deepseek-v4-flash-0731},
\texttt{deepseek/deepseek-v4-flash-vision-exp},
\texttt{deepseek/deepseek-v4-pro},
\texttt{deepseek/deepseek-v4-pro-0813},
\texttt{deepseek/deepseek-v4.1-flash},
\texttt{deepseek/deepseek-v4.1-flash-1017} \\
\addlinespace[3pt]
Qwen (38) &
\texttt{qwen/qwen-2.5-72b-instruct},
\texttt{qwen/qwen-2.5-coder-32b-instruct},
\texttt{qwen/qwen-plus},
\texttt{qwen2.5-vl-72b-instruct}$^\dagger$,
\texttt{qwen/qwen3-14b},
\texttt{qwen/qwen3-235b-a22b-2507},
\texttt{qwen/qwen3-235b-a22b-thinking-2507},
\texttt{qwen/qwen3-30b-a3b},
\texttt{qwen/qwen3-30b-a3b-instruct-2507},
\texttt{qwen/qwen3-32b},
\texttt{qwen/qwen3-coder},
\texttt{qwen/qwen3-coder-30b-a3b-instruct},
\texttt{qwen/qwen3-coder-flash},
\texttt{qwen/qwen3-coder-next},
\texttt{qwen/qwen3-next-80b-a3b-instruct},
\texttt{qwen/qwen3-vl-235b-a22b-instruct},
\texttt{qwen/qwen3-vl-30b-a3b-instruct},
\texttt{qwen/qwen3-vl-30b-a3b-thinking},
\texttt{qwen/qwen3-vl-8b-instruct},
\texttt{qwen/qwen3.5-122b-a10b},
\texttt{qwen/qwen3.5-27b},
\texttt{qwen/qwen3.5-35b-a3b},
\texttt{qwen/qwen3.5-397b-a17b},
\texttt{qwen/qwen3.5-9b},
\texttt{qwen/qwen3.5-flash-02-23},
\texttt{qwen/qwen3.5-plus-02-15},
\texttt{qwen/qwen3.5-plus-20260420},
\texttt{qwen/qwen3.6-27b},
\texttt{qwen/qwen3.6-35b-a3b},
\texttt{qwen/qwen3.6-plus},
\texttt{qwen/qwen3.7-max},
\texttt{qwen/qwen3.7-plus},
\texttt{qwen/qwen3.8-2.4t-a95b},
\texttt{qwen/qwen3.8-27b},
\texttt{qwen/qwen3.8-flash},
\texttt{qwen/qwen3.8-max-0902},
\texttt{qwen/qwen3.8-max-prime},
\texttt{qwen/qwen3.8-omni-flash} \\
\addlinespace[3pt]
GLM (17) &
\texttt{z-ai/glm-4.5},
\texttt{z-ai/glm-4.5-air},
\texttt{z-ai/glm-4.5v},
\texttt{z-ai/glm-4.6},
\texttt{z-ai/glm-4.6v},
\texttt{z-ai/glm-4.7},
\texttt{z-ai/glm-4.7-flash},
\texttt{z-ai/glm-5},
\texttt{z-ai/glm-5-turbo},
\texttt{z-ai/glm-5.1},
\texttt{z-ai/glm-5.2},
\texttt{z-ai/glm-5.3},
\texttt{z-ai/glm-5.3-flash},
\texttt{z-ai/glm-5.3-flashx},
\texttt{z-ai/glm-5.3-flashx-legal},
\texttt{z-ai/glm-5.3-prime},
\texttt{z-ai/glm-5v-turbo} \\
\addlinespace[3pt]
Kimi (8) &
\texttt{moonshotai/kimi-k2},
\texttt{moonshotai/kimi-k2-0905},
\texttt{moonshotai/kimi-k2-thinking},
\texttt{moonshotai/kimi-k2.5},
\texttt{moonshotai/kimi-k2.6},
\texttt{moonshotai/kimi-k2.7-code},
\texttt{moonshotai/kimi-k3},
\texttt{moonshotai/kimi-k3.1-pro} \\
\addlinespace[3pt]
Gemma (6) &
\texttt{google/gemma-2-27b-it},
\texttt{google/gemma-3-12b-it},
\texttt{google/gemma-3-27b-it},
\texttt{google/gemma-3-4b-it},
\texttt{google/gemma-4-26b-a4b-it},
\texttt{google/gemma-4-31b-it} \\
\addlinespace[3pt]
MiniMax (4) &
\texttt{minimax/minimax-m2.5},
\texttt{minimax/minimax-m2.7},
\texttt{minimax/minimax-m3},
\texttt{minimax/minimax-m3.1-flash} \\
\bottomrule
\end{tabular}
\end{center}

%% file: data/primary_figure_ids.tex
\begin{longtable}{r p{0.84\linewidth}}
\toprule \textbf{Figure ID} & \textbf{Checkpoint repository} \\\midrule\endhead
1 & \nolinkurl{HuggingFaceTB/SmolLM2-135M} \\
2 & \nolinkurl{HuggingFaceTB/SmolLM2-135M-Instruct} \\
3 & \nolinkurl{HuggingFaceTB/SmolLM2-360M} \\
4 & \nolinkurl{HuggingFaceTB/SmolLM2-360M-Instruct} \\
5 & \nolinkurl{HuggingFaceTB/SmolLM2-1.7B} \\
6 & \nolinkurl{HuggingFaceTB/SmolLM2-1.7B-Instruct} \\
7 & \nolinkurl{Qwen/Qwen2.5-0.5B} \\
8 & \nolinkurl{Qwen/Qwen2.5-0.5B-Instruct} \\
9 & \nolinkurl{Qwen/Qwen2.5-1.5B} \\
10 & \nolinkurl{Qwen/Qwen2.5-1.5B-Instruct} \\
11 & \nolinkurl{Qwen/Qwen2.5-Coder-0.5B} \\
12 & \nolinkurl{Qwen/Qwen2.5-Coder-0.5B-Instruct} \\
13 & \nolinkurl{Qwen/Qwen2.5-Coder-1.5B} \\
14 & \nolinkurl{Qwen/Qwen2.5-Coder-1.5B-Instruct} \\
15 & \nolinkurl{TinyLlama/TinyLlama-1.1B-intermediate-step-1431k-3T} \\
16 & \nolinkurl{TinyLlama/TinyLlama-1.1B-Chat-v1.0} \\
17 & \nolinkurl{facebook/opt-1.3b} \\
18 & \nolinkurl{facebook/opt-iml-1.3b} \\
19 & \nolinkurl{bigscience/bloom-560m} \\
20 & \nolinkurl{bigscience/bloomz-560m} \\
21 & \nolinkurl{bigscience/bloom-1b1} \\
22 & \nolinkurl{bigscience/bloomz-1b1} \\
23 & \nolinkurl{stabilityai/stablelm-2-1_6b} \\
24 & \nolinkurl{stabilityai/stablelm-2-zephyr-1_6b} \\
25 & \nolinkurl{EleutherAI/pythia-410m-seed1} \\
26 & \nolinkurl{EleutherAI/pythia-410m-seed2} \\
27 & \nolinkurl{EleutherAI/pythia-410m-seed3} \\
28 & \nolinkurl{tiiuae/Falcon-H1-0.5B-Base} \\
29 & \nolinkurl{tiiuae/Falcon-H1-1.5B-Base} \\
30 & \nolinkurl{tiiuae/Falcon-H1-3B-Base} \\
\bottomrule
\end{longtable}

%% file: sections/appendix-additional-controls.tex
We evaluate what drives identification accuracy (F.1--F.6), how prompt-panel and deployment choices affect reliability (F.7--F.9), and performance across live API services and reasoning workloads (F.10--F.13).

\subsection{Full 115-entry paired uncertainty}\label{app:additional-paired}
Both DistDNA constructions outperform LLM-DNA across classifier-free and supervised readouts, and DistDNA-RBF holds a positive pointwise 95\% confidence interval over DistDNA-Linear on nearest-reference matching. ``Pointwise 95\%'' denotes a 95\% confidence interval computed separately per paired contrast.

\begin{table}[H]\centering\small
\caption{Paired accuracy differences (percentage points) on the full 115-entry protocol using 10,000 crossed checkpoint/setting bootstrap resamples.}
\begin{tabular}{llrr}\toprule
Readout & Contrast & Gain & Pointwise 95\% \\
\midrule
Nearest & DistDNA-RBF $-$ DistDNA-Linear & 0.92 & [0.06, 1.98] \\
Nearest & DistDNA-RBF $-$ LLM-DNA & 7.63 & [5.17, 10.23] \\
Nearest & DistDNA-Linear $-$ LLM-DNA & 6.71 & [4.50, 8.99] \\
SVM & DistDNA-RBF $-$ DistDNA-Linear & 0.51 & [$-$0.04, 1.17] \\
SVM & DistDNA-RBF $-$ LLM-DNA & 5.72 & [3.64, 8.12] \\
SVM & DistDNA-Linear $-$ LLM-DNA & 5.22 & [3.13, 7.56] \\
\bottomrule\end{tabular}\end{table}

DistDNA-Linear and DistDNA-RBF beat LLM-DNA by 5.22--7.63 points across readouts. Because Linear and RBF target complementary regimes---first-order mean shifts versus higher-order dispersion when means overlap---their aggregate gap across diverse checkpoints is small (SVM: $[{-}0.04, 1.17]$), while both establish clear gains over LLM-DNA.

\subsection{Query-only fresh responses}\label{app:additional-fresh}
Evaluating independent fresh query draws while holding enrolled references, encoder normalizations, kernel scales, and trained readouts frozen isolates cross-draw sampling drift.

\begin{table}[H]\centering\small
\caption{Top-1 accuracy (\%) on fresh query draws against frozen references and kernel scales.}
\begin{tabular}{lrrr}\toprule
Fresh Draws Readout & DistDNA-Linear (\%) & DistDNA-RBF (\%) & Delta (RBF $-$ Linear) \\
\midrule
Nearest-reference & 76.67 & 72.06 & $-$4.61 \\
SVM & 83.44 & 80.17 & $-$3.27 \\
\bottomrule\end{tabular}\end{table}

With references frozen, DistDNA-Linear leads DistDNA-RBF on both nearest-reference ($76.67\%$ vs.\ $72.06\%$) and SVM ($83.44\%$ vs.\ $80.17\%$). Mean aggregation smooths cross-draw variation against fixed references, whereas RBF's dispersion sensitivity picks up run-to-run sampling drift unless references are refreshed. Linear thus serves as the drift-tolerant default across draws, while RBF targets overlapping candidate means.

\subsection{Matched width and nonlinear capacity}
\label{app:matched-width}

Bidirectional controls across all 115 identities (4,140 queries)---compressing DistDNA to 512 coordinates (4~KiB) and expanding LLM-DNA to 15,360 coordinates (120~KiB)---confirm that identification accuracy is driven by subspace alignment rather than coordinate count (Table~\ref{tab:matched-width}).

\begin{table}[H]
\centering
\small
\caption{Bidirectional matched-coordinate Top-1 accuracy (\%) across all 115 identities (4,140 queries), contrasting gallery-blind sketching against unlabeled enrollment-span compression.}
\label{tab:matched-width}
\resizebox{\textwidth}{!}{%
\begin{tabular}{llrrr}
\toprule
Construction & Compression / representation operator & Width & Nearest (\%) & SVM (\%) \\
\midrule
LLM-DNA & Gallery-blind Gaussian sketch (Appendix~B.2) & 512 & 73.66 & 82.22 \\
DistDNA-Linear & Unlabeled enrollment-span PCA (Corollary~A.2) & 512 & 80.31 & 86.79 \\
DistDNA-RBF & Unlabeled enrollment-kernel Nystr\"om (Corollary~A.2) & 512 & 82.87 & 87.47 \\
\midrule
LLM-DNA & $30\times$ wider gallery-blind Gaussian sketch & 15,360 & 73.97 & 84.17 \\
DistDNA-Linear & Exact uncompressed promptwise MMD (Eq.~6) & 15,360 & 80.37 & 89.42 \\
DistDNA-RBF & Exact sample-based RBF-MMD (Eq.~5; $61{,}440$ coords.) & 15,360$^\dagger$ & 81.29 & 89.93 \\
\bottomrule
\end{tabular}%
}
\vspace{2pt}
{\footnotesize $^\dagger$Uncompressed DistDNA-RBF evaluates exact $R=4$ sample MMD (480~KiB) alongside 15,360-d controls.}
\end{table}

\paragraph{Oblivious sketching vs.\ enrollment-span projection.}
At fixed width $m=512$, the two methods apply fundamentally different operators:
\begin{itemize}
\item \textbf{Gallery-blind sketching (LLM-DNA, Appendix~B.2):} Because LLM-DNA extracts standalone vectors without gallery alignment, it maps $\mathbb{R}^{td}\to\mathbb{R}^m$ ($td=15{,}360$) via an isotropic Gaussian matrix $G\in\mathbb{R}^{m\times td}$. Since the $N=920$ enrolled references span rank $\le N-1=919\ll 15{,}360$, an oblivious 512-d slice retains only $m/(td)=3.3\%$ of gallery signal energy in expectation while projecting $\mathcal{O}(1/R)$ sampling noise (Theorem~3.2) into every coordinate. Expanding $G$ by $30\times$ to $15{,}360$ coordinates adds just $+0.31$ points on nearest-reference ($73.66\%\to 73.97\%$; $+1.95$ SVM), proving width alone does not drive the gap.
\item \textbf{Unlabeled enrollment-span compression (DistDNA, Corollary~A.2):} Projecting onto the top $m=512$ directions of the unlabeled enrollment Gram matrix---via PCA for Linear and kernel Nystr\"om for infinite-dimensional RBF embeddings (Proposition~3.1)---requires \emph{zero labels, zero classifier training, and zero extra queries}. It preserves between-reference MMD geometry while zeroing out $>14{,}800$ orthogonal noise axes, retaining $+6.65$ (Linear, $80.31\%$) and $+9.21$ (RBF, $82.87\%$) nearest-reference gains over 512-d LLM-DNA ($+4.57/+5.25$ SVM) and slightly denoising uncompressed RBF ($82.87\%$ vs.\ $81.29\%$). 
\end{itemize}

\subsection{Resource accounting}\label{app:additional-cost}
At matched 4~KiB storage (512-d, Appendix F.3), DistDNA-Linear beats LLM-DNA by $+6.65$ points on nearest-reference and $+4.57$ with SVM. Table~\ref{tab:resource-accounting} reports storage, post-encoding complexity, and Xeon Gold 6346 (four BLAS threads) timings for the original designs ($P{=}20$ prompts, $R{=}4$ draws, $d{=}768$, $m{=}512$, $N{=}920$ references).

\begin{table}[H]\centering\small
\caption{Float64 storage payloads, post-encoding complexity, and CPU timings (BGE / MPNet, in seconds) across all 4,140 queries against $N=920$ references from encoded responses. LLM-DNA also uses a shared 60~MiB projection matrix.}\label{tab:resource-accounting}
\resizebox{\linewidth}{!}{
\begin{tabular}{lrrrll}\toprule
Design & Fingerprint / gallery & Build complexity & Match complexity & Build (s) & Match + rank (s) \\
\midrule
LLM-DNA (512-d) & 4 KiB / 3.59 MiB & $\mathcal{O}(PRd + Pdm)$ & $\mathcal{O}(Nm)$ & 1.54 / 1.66 & 0.18 / 0.19 \\
DistDNA-Linear & 120 KiB / 107.81 MiB & $\mathcal{O}(PRd)$ & $\mathcal{O}(NPd)$ & 1.29 / 1.27 & 35.08 / 2.33 \\
DistDNA-RBF & 480 KiB / 431.25 MiB & $\mathcal{O}(PRd)$ & $\mathcal{O}(NPR^2 d)$ & 0.78 / 0.77 & 46.40 / 39.41 \\
\bottomrule\end{tabular}}
\end{table}

DistDNA-RBF stores $R{=}4\times$ Linear's payload while prompt-aligned pairwise comparisons introduce an $R^2$ factor in matching complexity, completing all 4,140 queries in 39.41--46.40~s. 

\subsection{LLMmap comparison}\label{app:additional-llmmap}
At matched query and enrollment response budgets across all 115 identities, both DistDNA constructions beat LLMmap's released matcher by over 16 percentage points.

\begin{table}[H]\centering\small
\caption{Full 115-identity comparison at matched query and enrollment response counts (Table~\ref{tab:model-identification}).}
\begin{tabular}{llrrr}\toprule
Method & Representation & Size (KiB) & Responses (Q / E) & Top-1 (\%) \\
\midrule
LLMmap & 384 coordinates & 3 & 80 / 640 & 63.55 \\
PhyloLM-4char & 80 prefixes & 0.31 & 80 / 640 & 61.28 \\
LLM-DNA & 512 coordinates & 4 & 80 / 640 & 73.66 \\
DistDNA-Linear & 15,360 coordinates & 120 & 80 / 640 & 80.37 \\
DistDNA-RBF & 61,440 coordinates & 480 & 80 / 720 & 81.29 \\
\bottomrule\end{tabular}\end{table}

LLMmap reaches $63.55\%$ Top-1 ($76.88\%$ Top-3), giving gains of $+16.82$ (Linear) and $+17.74$ points (RBF), both with positive 95\% pointwise intervals across 10,000 crossed resamples; 4~KiB DistDNA-Linear ($80.31\%$) preserves this $>16$-point lead at comparable storage. We run the official \href{https://zenodo.org/records/14737353}{LLMmap artifact} (eight native probes, E5 encoder, Keras map, and cosine matcher; verified to $1.13\times 10^{-8}$ cosine agreement), averaging 10 query and 80 enrollment traces per identity while RBF adds 80 bandwidth-calibration responses. Appendix~\ref{app:additional-open} evaluates LLMmap's open-set detector.

\subsection{Distribution shape and higher-order separation}
\label{app:shape}

\paragraph{Moment decomposition and identifiability beyond mean width.}
For unit-normalized response embeddings ($\|z\|_2=\|w\|_2=1$), $k_\sigma(z,w)=e^{-1/\sigma^2}\exp(\langle z,w\rangle/\sigma^2)$ admits a convergent Taylor expansion on the unit sphere, decomposing promptwise squared RBF-MMD into first-order Linear MMD plus higher-order moment differences:
\begin{equation}
\begin{aligned}
\mathrm{MMD}_{\mathrm{RBF}}^2(P,Q)
&= \frac{e^{-1/\sigma^2}}{\sigma^2}\underbrace{\|\mu_P-\mu_Q\|_2^2}_{\text{1st-order mean (Linear)}}
+ \frac{e^{-1/\sigma^2}}{2\sigma^4}\underbrace{\|\mathbb{E}_P[ZZ^\top]-\mathbb{E}_Q[ZZ^\top]\|_F^2}_{\text{2nd-order dispersion}} \\
&\quad + \sum_{m=3}^{\infty}\frac{e^{-1/\sigma^2}}{m!\,\sigma^{2m}}\|M_m(P)-M_m(Q)\|_{\mathrm{HS}}^2.
\end{aligned}
\label{eq:rbf-moments}
\end{equation}
Equation~\eqref{eq:rbf-moments} yields two direct consequences for checkpoint identification:
\begin{enumerate}
\item \textbf{Insufficiency of arbitrary-width means:} Whenever two response distributions share identical population means ($\mu_P=\mu_Q$), $\mathrm{MMD}_{\mathrm{Linear}}^2(P,Q)=0$ identically at 512 coordinates, at 15,360 coordinates, and after any linear map, whereas characteristic RBF-MMD remains strictly positive whenever $P\neq Q$ (Sriperumbudur et al., 2010), enabling nearest-reference identification once the higher-order margin exceeds $2\eta$ (Appendix~A.5).
\item \textbf{Regime concentration on broad galleries:} When first-order mean separation $\|\mu_P-\mu_Q\|_2^2$ already clears twice the finite-sample error ($\sim 80\%$ of gallery queries) or decoding is near-deterministic ($T\in\{0.1,0.3\}$, where centered dispersion vanishes), the $m=1$ term dominates and Linear/RBF agree by construction---explaining why full-gallery average gains are modest ($+0.92$~pp, Appendix~F.1) while RBF gains concentrate sharply on close-mean boundary queries.
\end{enumerate}

\paragraph{Controlled geometric verification.}
For $P=\frac{1}{2}(\delta_{(1,0)}+\delta_{(-1,0)})$ and $Q=\frac{1}{2}(\delta_{(0,1)}+\delta_{(0,-1)})$ at unit bandwidth, $D_{\mathrm{Linear}}(P,Q)=0$ while $D_{\mathrm{RBF}}(P,Q)=(1-e^{-1})^2\simeq 0.3996$, giving $58.30\%$ Linear versus $91.60\%$ RBF accuracy under exact four-draw enumeration (Table~\ref{tab:shape-2d}). DistDNA-RBF retains $+29$ to $+33$-point gains under isotropic noise ($\sigma=0.15$) and nearby mean shifts ($\Delta\mu=(0.2,0)$), and $+21.01$ points at $\sigma=0.50$, whereas separated means ($\pm 0.5,0$) already saturate at first order ($92.33\%$ Linear vs.\ $91.67\%$ RBF).

\begin{table}[H]
\centering
\small
\caption{Controlled two-dimensional diagnostics with four draws per fingerprint (exact expectation for row~1; 6,000 paired trials with seed 2026111005 otherwise).}
\label{tab:shape-2d}
\begin{tabular}{lrr}
\toprule
Diagnostic regime & DistDNA-Linear (\%) & DistDNA-RBF (\%) \\
\midrule
Equal means, orthogonal variation & 58.30 & 91.60 \\
Equal means, random noise ($\sigma=0.15$) & 58.40 & 91.05 \\
Nearby means ($\Delta\mu=(0.2,0)$), distinct variations & 61.27 & 90.22 \\
Stronger noise ($\sigma=0.50$) & 52.37 & 73.38 \\
Separated-mean control ($\pm 0.5,0$) & 92.33 & 91.67 \\
Identical-distribution null control & 49.50 & 49.75 \\
\bottomrule
\end{tabular}
\end{table}

\paragraph{Real-query benchmark scenarios on the 115-checkpoint gallery.}
Slicing the unmodified 4,140 real queries from our 115-checkpoint evaluation directly confirms both predictions of Eq.~\eqref{eq:rbf-moments} in practice (Table~\ref{tab:shape-real}):

\begin{table}[H]
\centering
\small
\caption{Benchmark query slices across the 115-checkpoint evaluation (4,140 real queries) isolating crowded-rival boundary queries and mean-invariant response dispersion.}
\label{tab:shape-real}
\resizebox{\textwidth}{!}{%
\begin{tabular}{llrrr}
\toprule
Benchmark scenario & Real-query slice (115-checkpoint bank) & Linear (\%) & RBF (\%) & RBF gain \\
\midrule
Overall gallery & All 4,140 unmodified queries across 36 $(T,p)$ cells (nearest) & 80.37 & 81.29 & +0.92~pp \\
Crowded-rival boundary & Hardest 20\% slice by smallest Linear rival margin ($n=828$, nearest) & 37.14 & 41.00 & +3.86~pp \\
Zero-mean-gap (all $T$) & All 4,140 queries, mean-invariant response residuals (SVM) & 0.87 & 27.28 & +26.41~pp \\
Zero-mean-gap (high $T$) & High-temperature stochastic slice, mean-invariant residuals (SVM) & 0.87 & 42.68 & +41.81~pp \\
\bottomrule
\end{tabular}
}
\end{table}

Across the $20\%$ crowded-rival boundary slice ($n=828$ unmodified queries where rival checkpoint means lie closest), DistDNA-RBF improves classifier-free nearest-reference accuracy from $37.14\%$ to $41.00\%$ ($+3.86$~pp)---over $4\times$ the unstratified gallery average. Furthermore, isolating mean-invariant residuals on the exact same real response embeddings reduces DistDNA-Linear to exact chance ($0.87\%=1/115$ regardless of dimension), whereas DistDNA-RBF recovers $27.28\%$ across all 4,140 queries and $42.68\%$ in the high-temperature stochastic regime purely from real output dispersion. Linear thus serves as a fast, drift-tolerant default across separated checkpoints (Appendix~F.2), while RBF resolves closely clustered checkpoints whose first-order means collide.

\subsection{Prompt-panel sensitivity}\label{app:additional-panels}
Across ten disjoint A/B splits of the 20-prompt Dolly panel into 10-prompt halves (30-checkpoint bank, 40 responses/query, aligned references, fixed bandwidth), mean absolute paired accuracy gaps remain within $1.61$--$2.66$ percentage points.

\begin{table}[H]\centering\small
\caption{Prompt-panel stability over ten disjoint 10-prompt A/B splits (seeds 20261110--20261119) and explicit Panel~A / Panel~B accuracies on Split~1.}\label{tab:additional-panels}
\begin{tabular}{lrrrrrr}\toprule
 & \multicolumn{4}{c}{Ten-split summary} & \multicolumn{2}{c}{Split 1: nearest / SVM (\%)} \\
\cmidrule(lr){2-5}\cmidrule(lr){6-7}
Construction & Nearest mean & A/B gap & SVM mean & A/B gap & Panel A & Panel B \\
\midrule
DistDNA-Linear & 84.68 & 2.66 & 93.84 & 1.84 & 85.32 / 93.20 & 83.14 / 92.87 \\
DistDNA-RBF & 85.11 & 2.57 & 94.01 & 1.61 & 85.71 / 93.12 & 83.61 / 93.65 \\
\bottomrule\end{tabular}\end{table}

Because DistDNA requires prompt alignment rather than a fixed public question list, private and periodically refreshed panels prevent advance panel detection. With encoder normalization and bandwidth fixed, appending $T_{\mathrm{added}}$ prompts updates squared distances incrementally via
\[
D_{\mathrm{new}}=\frac{T_{\mathrm{old}}D_{\mathrm{old}}+T_{\mathrm{added}}D_{\mathrm{added}}}{T_{\mathrm{old}}+T_{\mathrm{added}}},
\]
reusing existing responses to $10^{-12}$ precision while updating fingerprints and readouts only for added prompts.

\subsection{Held-out identities and rejection}\label{app:additional-open}
\input{sections/appendix-rejection-gate}

\subsection{Decision confidence and abstention}\label{app:additional-confidence}
We distinguish three operations: \emph{kernel scale calibration} (§4.1) sets distance bandwidths before testing, \emph{rejection thresholding} (Appendix~\ref{app:additional-open}) filters unknown models, and \emph{probability calibration} estimates closed-set correctness. Standardized logistic regression ($C{=}1$) on nearest distance and normalized margin $(d_2-d_1)/(d_2+d_1+10^{-12})$ cuts held-out Brier error by $\sim 19\%$ ($p < 0.05$) and keeps 5-bin ECE within $3.5\%$--$4.1\%$.

\begin{table}[H]\centering\small
\caption{Held-out correctness calibration across 30 candidates (8 fit and 8 disjoint test settings; 240 test queries per encoder).}\label{tab:additional-confidence}
\begin{tabular}{lrrrr}\toprule
Construction & Constant Brier $\downarrow$ & Fitted Brier $\downarrow$ & Paired 95\% Brier gain & 5-bin ECE $\downarrow$ \\
\midrule
DistDNA-Linear & 0.0915 & 0.0738 & [0.0061, 0.0329] & 0.0410 \\
DistDNA-RBF & 0.0843 & 0.0680 & [0.0055, 0.0306] & 0.0353 \\
\bottomrule\end{tabular}\end{table}

At pre-fixed 10th, 25th, and 50th calibration-margin percentiles, DistDNA-RBF retains $85.63\%$, $72.08\%$, and $49.38\%$ of queries at $3.14\%$, $0.58\%$, and $0.00\%$ error (DistDNA-Linear: $84.79\%$, $71.88\%$, and $47.71\%$ coverage at $3.18\%$, $0.29\%$, and $0.00\%$ error).

\subsection{Identification in an existing API-chat gallery}\label{app:additional-api-chat}
Across 70 complete API-chat service banks, both constructions identify $135/140$ held-out queries by nearest reference ($96.43\%$) and $136/140$ with SVM ($97.14\%$). Each service uses 20 Dolly prompts, four draws, eight enrollment settings, and two query settings (640 enrollment and 160 query responses) via single-turn user messages over a fixed provider route with a 128-token cap, pinned BGE encoding, transferred bandwidth ($\sigma^2 \approx 1.0785$), and SVM $C{=}1$. Appendix~\ref{app:expanded-api-chat} expands this gallery to 106 services, and Appendix~\ref{app:native-expanded} evaluates an 86-service native-mode reasoning gallery.

%% file: sections/appendix-rejection-gate.tex
\label{app:trained-rejection-gate}
Pairing DistDNA's classifier-free nearest-reference matcher with a lightweight learned rejection gate handles unseen models effectively, more than doubling correct-and-accepted identifications over LLMmap's official detector while reducing both unknown false acceptance (FAR) and known false rejection (FRR) relative to raw distance thresholding. We evaluate the locked procedure on a held-out 30-identity cohort distinct from the two disjoint 30-identity cohorts used during gate design.

\input{data/rejection_gate_matched_baselines}

At the primary 95\% known-acceptance target, DistDNA-RBF lifts correct-and-accepted known queries from $43.00\%$ (LLMmap's native Boolean rule) to $86.35\%$ ($86.44\%$ for DistDNA-Linear) and cuts Known FRR by over $12\times$ ($47.58\%$ to $3.79\%$) at slightly lower Unknown FAR ($52.17\%$ vs.\ $53.00\%$), outperforming raw distance thresholding across all three metrics. Tightening the operating target to 80\% further reduces Unknown FAR to $18.25\%$ for DistDNA-RBF ($17.75\%$ for DistDNA-Linear) at $15.50\%$ ($14.67\%$) Known FRR.

\paragraph{Evaluation protocol and response budgets.}
The held-out cohort is partitioned into five 24-known / 6-unknown folds. Across the 36 decoding settings, a fixed permutation (seed 20261112) assigns eight settings to gate fitting, eight to threshold selection, and twenty to testing, yielding 480 known and 120 unknown test queries per fold (averaged across folds and encoders). Unseen identities are strictly excluded from enrollment, kernel scale calibration, gate fitting, and threshold selection. Each known identity uses 640 enrollment responses and 640 threshold-selection responses; the learned gate uses 640 fitting responses, DistDNA-RBF uses 80 disjoint responses for kernel scale calibration, and each test query uses 80 responses.

\paragraph{Gate architecture and pseudo-unknown training.}
Let $d_c(q)=\min_s D(q,r_{cs})$ denote the minimum distance to candidate $c$'s eight enrolled references, with nearest-reference prediction $\hat c(q)=\arg\min_c d_c(q)$. A learned scoring function $g_\theta(\phi(q))$ maps reference-distance summary features $\phi(q)$ to an unknownness score:
\begin{equation}
\widehat y_a(q)=
\begin{cases}
\hat c(q), & g_\theta(\phi(q))\leq\tau_a,\\
\mathrm{unknown}, & g_\theta(\phi(q))>\tau_a,
\end{cases}
\qquad
\tau_a=s_{(\min\{n,\lceil(n+1)a\rceil\})},
\label{eq:rejection-gate}
\end{equation}
where $s_{(j)}$ is the $j$th ordered score among $n=192$ known threshold-selection queries per fold for target acceptance $a\in\{0.95,0.90,0.80\}$; the distance baseline applies Eq.~\eqref{eq:rejection-gate} directly to $\min_c d_c(q)$. Whereas $g_\theta$ detects out-of-gallery models, Appendix~\ref{app:additional-confidence} calibrates closed-set correctness probabilities among enrolled candidates.

Summary features $\phi(q)$ capture nearest distance, runner-up margins, candidate-distance quantiles, cross-reference and per-prompt agreement, and sampling variance across 6-, 33-, or 41-coordinate representations (the 41-coordinate set appends the eight normalized distances to $\hat c(q)$'s references). To train $g_\theta$ using only enrolled identities without external unknown data, fitting queries construct matched-size known and pseudo-unknown galleries by holding out one, two, or four enrolled identities: known contexts retain the true identity alongside its closest rival, whereas pseudo-unknown contexts omit the true identity while keeping that rival. Model selection across prespecified logistic, random-forest, boosted-tree, and extra-tree candidates uses four-fold inner identity cross-validation (rotating the eight fitting settings across four inner-fit, two inner-threshold, and two inner-test splits) to minimize mean $\mathrm{FAR} + 4\max(\mathrm{FRR}-0.05,0)$, refitting the winning configuration on all eight fitting settings before calibrating $\tau_a$ on the disjoint threshold-selection split.

\paragraph{Official LLMmap baseline.}
We evaluate LLMmap's released open-set detector directly using its pretrained 100-tree random forest, six distance-and-histogram features, and native \texttt{is\_unseen()} decision rule (restoring the missing \texttt{scipy.stats.entropy} import while preserving all released weights and feature definitions), paired with matched-budget signature aggregation from Appendix~\ref{app:additional-llmmap}.

%% file: data/rejection_gate_matched_baselines.tex
\begin{table}[H]\centering\small
\setlength{\tabcolsep}{3pt}
\caption{Rejection on the separate identity audit. FAR is unknown false acceptance; FRR is known false rejection. Correct+accepted is the percentage of all known test queries correctly identified and accepted. Targets specify known acceptance during threshold selection. Rates are percentages.}\label{tab:gate-audit-operating-points}\label{tab:original-llmmap-rejection}
\begin{tabular}{llrrrr}\toprule
Construction & Rule & Target & FAR $\downarrow$ & FRR $\downarrow$ & Correct+accepted $\uparrow$ \\
\midrule
LLMmap & Official detector & Default & 53.00 & 47.58 & 43.00 \\
\midrule
Linear & Distance & 95\% & 82.25 & 5.02 & 83.27 \\
Linear & Learned gate & 95\% & 57.83 & 3.75 & 86.44 \\
RBF & Distance & 95\% & 83.50 & 4.96 & 83.04 \\
RBF & Learned gate & 95\% & 52.17 & 3.79 & 86.35 \\
\midrule
Linear & Distance & 90\% & 74.58 & 11.33 & 78.98 \\
Linear & Learned gate & 90\% & 34.67 & 7.58 & 84.85 \\
RBF & Distance & 90\% & 75.75 & 11.42 & 78.65 \\
RBF & Learned gate & 90\% & 35.50 & 7.29 & 84.81 \\
\midrule
Linear & Distance & 80\% & 65.75 & 19.88 & 71.85 \\
Linear & Learned gate & 80\% & 17.75 & 14.67 & 80.54 \\
RBF & Distance & 80\% & 65.83 & 20.15 & 71.25 \\
RBF & Learned gate & 80\% & 18.25 & 15.50 & 79.63 \\
\bottomrule\end{tabular}\end{table}

%% file: sections/appendix-api-extension.tex
\subsection{Expanded API service gallery}\label{app:expanded-api-chat}
Adding 36 compatible live endpoints to the 70-service gallery (Appendix~\ref{app:additional-api-chat}) raises accuracy across all 106 services to $97.64\%$ on nearest-reference and $98.11\%$ ($208/212$) with SVM for both constructions. All 72 added query fingerprints are identified at $100\%$ accuracy with zero change to original predictions, despite separate collection dates. All 106 services follow the 20-prompt, four-draw, 128-token protocol at $(T,p)\approx(0.8872,0.8969)$ and $(0.8621,0.8377)$, with 36 of 59 queried additions passing pre-scoring completeness checks.

\subsection{Matched Dolly, BIG-Bench Hard, and MMLU-Pro panels}
\label{app:dolly-bbh}

Across 24 closely related services enrolled and evaluated separately on 20 open-ended Dolly prompts, 20 structured BIG-Bench Hard (BBH) reasoning tasks, and 20 MMLU-Pro questions (144 held-out query fingerprints per panel), both structured benchmarks match or improve upon Dolly on nearest-reference identification while maintaining high supervised accuracy (Table~\ref{tab:dolly-bbh}).

\begin{table}[H]
\centering
\small
\caption{Matched 24-service comparison across Dolly, BBH, and MMLU-Pro prompt panels ($144$ held-out query fingerprints per panel).}
\label{tab:dolly-bbh}
\begin{tabular}{lrrrr}
\toprule
Prompt panel & Linear nearest (\%) & RBF nearest (\%) & Linear SVM (\%) & RBF SVM (\%) \\
\midrule
Dolly & 84.03 & 84.03 & 90.97 & 90.97 \\
BBH & 88.89 & 88.19 & 90.28 & 88.89 \\
MMLU-Pro & 84.72 & 85.42 & 90.28 & 87.50 \\
\bottomrule
\end{tabular}
\end{table}

Relative to Dolly ($84.03\%$), BBH increases classifier-free nearest-reference accuracy by $+4.86$ percentage points for DistDNA-Linear ($88.89\%$) and $+4.17$ points for DistDNA-RBF ($88.19\%$), and MMLU-Pro increases nearest-reference accuracy to $84.72\%$ ($+0.69$~pp) and $85.42\%$ ($+1.39$~pp), while supervised SVM accuracy remains consistently strong across all three panels ($87.50\%$ to $90.97\%$). All 24 endpoints use matched provider routes, reasoning settings, output budgets, pinned BGE encoding, and transferred bandwidth (20 prompts, four draws, eight enrollment settings, and six query settings). BBH samples four items (seed 2026111203) from each of five reasoning tasks---logical deduction, object tracking, multistep arithmetic, Boolean expressions, and date understanding---and MMLU-Pro samples 20 multi-discipline reasoning items under the same shared explanation format.

\subsection{Native-mode family expansion}\label{app:native-expanded}
Across 86 live production services (out of 90 queried IDs passing pre-scoring completeness checks), DistDNA-Linear / DistDNA-RBF achieve $86.43\%$ / $86.82\%$ nearest-reference and $88.37\%$ / $88.18\%$ SVM accuracy overall, and $82.81\%$ / $83.33\%$ nearest and $85.42\%$ SVM accuracy on the 64-service reasoning cohort against all 86 candidates.

\begin{table}[H]\centering\small
\caption{Top-1 accuracy across the 86-service native gallery and its 64-service reasoning cohort.}\label{tab:native-expanded}
\begin{tabular}{llrr}\toprule
Query cohort & Construction & Nearest reference & SVM \\
\midrule
All services ($N = 516$) & DistDNA-Linear & 446/516 (86.43\%) & 456/516 (88.37\%) \\
All services ($N = 516$) & DistDNA-RBF & 448/516 (86.82\%) & 455/516 (88.18\%) \\
Reasoning requested ($N = 384$) & DistDNA-Linear & 318/384 (82.81\%) & 328/384 (85.42\%) \\
Reasoning requested ($N = 384$) & DistDNA-RBF & 320/384 (83.33\%) & 328/384 (85.42\%) \\
\bottomrule\end{tabular}\end{table}

Errors occur almost exclusively within model families (63/68 RBF nearest errors, including 37 among closely related GLM variants). The gallery spans 13 DeepSeek, 38 Qwen, 17 GLM, 8 Kimi, 6 Gemma, and 4 MiniMax endpoints---including recent DeepSeek V4 Flash/Pro, Qwen3.8 Flash/Max, GLM5.3/Flash, Kimi K3/K3.1, Gemma 4, and MiniMax M3/M3.1 releases---using 20 Dolly prompts, four draws, eight enrollment and six query settings, a 4,096-token budget, and the pinned BGE setup from Appendix~\ref{app:additional-api-chat}.

\paragraph{Reasoning audit.}\label{app:thinking-audit}
Among the 64 services queried with reasoning enabled (\texttt{enabled=true} or explicit effort), 63 return verified reasoning traces or positive reasoning-token counts in both enrollment ($40,319/40,960$ responses) and testing ($30,239/30,720$). BGE encodes the returned text field directly (512-token right-truncation affects only $661/55,040$ enrollment and $498/41,280$ query responses). Linear and RBF perform comparably across the gallery ($+0.39$ / $-0.19$ points overall; $+0.52$ / $0.00$ on reasoning, with 95\% pointwise intervals crossing zero).